\documentclass[manuscript,screen,natbib=false]{acmart}
\setcopyright{none}
\renewcommand\footnotetextcopyrightpermission[1]{}
\usepackage{graphicx,booktabs,longtable,array,pdflscape,float}
\usepackage{amsmath}
\usepackage[backend=biber,style=numeric,sorting=none,maxbibnames=99]{biblatex}
\usepackage{microtype,enumitem,xcolor,colortbl,placeins,needspace}
\newcommand{\ab}[1]{{\color{black}#1}}
\title{Signals, Evidence, and Context: User Scrutiny in Real-World LLM Conversations}
\author{Ahana Biswas}
\affiliation{%
  \institution{School of Computing and Information, University of Pittsburgh}
  \city{Pittsburgh}
  \state{Pennsylvania}
  \country{USA}
}
\author{Yu-Ru Lin}
\affiliation{%
  \institution{School of Computing and Information, University of Pittsburgh}
  \city{Pittsburgh}
  \state{Pennsylvania}
  \country{USA}
}

\hypersetup{pdfauthor={Ahana Biswas and Yu-Ru Lin}}
\begin{abstract}
\ab{Large language models (LLMs) often present synthesized answers without making their supporting evidence easy to inspect. We study how features of these answers relate to users' subsequent accepting, checking, and challenging in 489,283 assistant--user response pairs from ShareChat and WildChat. Broad attribution language appears in 31.9\% of qualifying responses, compared with explicit URLs in 1.33\%, revealing a gap between claims of grounding and preserved inspectable support. Attribution, uncertainty, and other response cues generally accompany both checking and challenging; only uncertainty shows a small statistically distinguishable pooled association favoring checking. Within Technical Help, uncertainty favors checking relative to challenging; refusals and reasoning traces favor challenging in Seeking Information. Scrutiny is concentrated in some conversations, while its apparent turn-to-turn persistence becomes much smaller when conversation-level differences and response ordering are examined separately. These results suggest that supporting verification requires attention to what evidence can be inspected, which question a user is trying to resolve, and the history of the interaction.}

\end{abstract}

\begin{document}
\maketitle
\raggedbottom
\setlength{\textfloatsep}{6pt plus 2pt minus 2pt}
\setlength{\intextsep}{6pt plus 2pt minus 2pt}
\setlength{\abovedisplayskip}{5pt plus 2pt minus 2pt}
\setlength{\belowdisplayskip}{5pt plus 2pt minus 2pt}
\setlength{\abovedisplayshortskip}{2pt plus 1pt minus 1pt}
\setlength{\belowdisplayshortskip}{3pt plus 1pt minus 1pt}

\section{Introduction}
\ab{An answer from a large language model (LLM) can invoke an official report or expert judgment without showing the evidence behind its claims. The consequences of relying on apparent authority are evident in \emph{Mata v. Avianca}, where attorneys submitted fictitious judicial authorities generated by an AI system and were sanctioned by a federal court \cite{mata2023sanctions}. That case involved fabricated sources. In everyday information seeking, a cited source can be genuine yet difficult for a user to identify or inspect.}

\ab{LLM interfaces complicate familiar credibility judgments. Conventional web search exposes documents that a user can compare, while a conversational assistant often presents a synthesized answer before exposing its sources. In everyday conversations, users may ask for evidence, dispute a response, or simply move on. How those choices relate to the evidence and cues in an assistant's answer remains poorly understood.}

\ab{Consider an assistant answering an application-deadline question: ``According to the agency's official notice, applications close Friday.'' Invoking an ``official notice'' is a \emph{verification signifier}, language suggesting authority without exposing the notice. A link to the relevant notice would be \emph{material verification support}, while offering to locate it or allowing the user to ask for it provides a \emph{verification pathway}. Uncertainty or confidence language communicates the assistant's epistemic stance. A user might ask ``Where does it say that?'' (\emph{checking}), reply ``The notice says Monday'' (\emph{challenging}), or proceed (\emph{accepting}). This is an illustration, not a corpus quotation. We call the configuration of supports, signifiers, epistemic cues, and available actions the \emph{verification environment}. Checking and challenging together constitute \emph{scrutiny}, an observable act whose correctness is not established by the transcript.}

\ab{A signifier can suggest that outside evidence exists, a link creates an opportunity to inspect it, and a request for confirmation can return the user to the same system that generated the claim. These differences matter across tasks: uncertainty may invite a diagnostic question in technical troubleshooting, while an unhelpful refusal may prompt disagreement during information seeking. Epistemic vigilance and information-foraging research suggests that checking depends on its value, cost, and stakes \cite{sperber2010,pirolli1999information,tankelevitch2024metacognitive}.}

\ab{We analyze 489,283 assistant--user response pairs from ShareChat and WildChat \cite{yan2025sharechat,zhao2024wildchat,ai2wildchat48m}. \textbf{RQ1} asks what verification-related features appear in these recorded responses and whether grounding language is accompanied by inspectable material. \textbf{RQ2} examines how those features relate to accepting, checking, and challenging at the next user turn. \textbf{RQ3} studies associations within tasks and corpora, since the same cue may be encountered in different activities and the two collections have different model mixes, collection pathways, and preserved source fields. \textbf{RQ4} asks whether scrutiny is concentrated in particular conversations or additionally recurs because of the order of exchanges. Together, the questions trace how information is presented, how users visibly respond, and how those responses are distributed across longer interactions.}

\ab{We connect response-level evidence and cues to observable user behavior across repeated exchanges. This empirical account informs evaluations of whether an assistant makes claims inspectable and whether subsequent requests for evidence or corrections remain unresolved.}

\section{Related Work and Conceptual Framework}
\subsection{Credibility judgments when an interface speaks for its sources}
Web credibility research locates judgments in the interaction of source identity, content, presentation, and task. Rieh distinguishes anticipatory decisions about what to consult from evaluation during use \cite{rieh2002}. Hilligoss and Rieh describe credibility assessment across an information-seeking encounter \cite{hilligoss2008}; Metzger and Flanagin show why users rely on heuristics when source evaluation is costly \cite{metzger2007,metzger2013credibility}. The MAIN model links technological affordances to those heuristics \cite{sundar2008main}, while lateral reading recommends leaving the focal page to examine independent evidence \cite{wineburg2019lateral}.

Conversational synthesis blurs the boundary between source and summary. The same assistant may generate a claim, name its purported authority, and describe the evidence supporting it. Work on attribution and generative search documents citations that are present without reliable claim-level support \cite{liu2023verifiability,gao2023alce,rashkin2023attribution}; qualitative research on answer engines identifies a similar gap between cited responses and verifiability \cite{venkit2024answerengines}. A source marker and evidence that supports the associated claim remain distinct objects of evaluation.

\subsection{Trust, reliance, and observable checking}
Human-factors research distinguishes trust, an attitude toward an agent under uncertainty, from reliance, a behavioral choice \cite{lee2004trust}. Trust can be dispositional, situational, or learned \cite{hoff2015trust}, and construct reviews warn against substituting observed dependence for trustworthiness \cite{kohn2021measurement}. Appropriate reliance additionally depends on the accuracy of the advice and the task the user faces \cite{schemmer2023appropriate,guo2024decision}. Conversational records document what users say next, while leaving their private beliefs and subsequent decisions largely unobserved.

Controlled studies establish that presentation can alter human responses to AI advice. Explanations can increase reliance without improving complementary decision performance \cite{bansal2021whole}. Cognitive forcing may curb overreliance at the cost of effort \cite{bucinca2021trust}, and the value of explanations depends on that effort \cite{vasconcelos2023explanations}. For LLMs, uncertainty language can change reliance \cite{kim2024notsure}; source access may help identify incorrect answers \cite{kim2025fostering}, but citations can also increase trust when support is weak \cite{ding2025citations}. Those experiments establish cue effects under controlled conditions, while observational conversations show the mixed environments in which users encounter them.

\subsection{Verification affordances, signifiers, and epistemic communication}
\ab{Affordances are possibilities for action supported by an environment, whereas signifiers communicate an apparent opportunity or status \cite{gibson1979,norman2013,hutchby2001}. In the running example, ``According to the agency's official notice, applications close Friday'' is a signifier; a linked agency notice is material support; ``Shall I check the notice?'' suggests a pathway. Uncertainty, confidence, disclaimers, and refusals communicate an epistemic stance or boundary. Linguistic markers of hedging are communicative practices, distinct from calibration \cite{hyland1998hedging,mielke2022linguistic}.}

Question-driven transparency research emphasizes the user's immediate information need \cite{liao2021question,liao2024transparency}. Recent studies observe whether conversational tools prompt source inspection and how people move between chat and search \cite{liao2026chatbutverify,mayerhofer2025blending}. In each case, what an interface makes inspectable matters as much as how credibly the answer is presented.

\subsection{Epistemic vigilance, costs, and temporal interaction}
Epistemic vigilance concerns selective assessment of communicators and claims \cite{sperber2010}. Information foraging models evidence seeking as a choice weighed against time and effort \cite{pirolli1999information}; work on AI-assisted metacognition similarly documents the effort of deciding what to delegate and monitor \cite{tankelevitch2024metacognitive}. The utility of another check will differ for familiar, consequential, or difficult-to-verify tasks.

Large conversation collections capture such heterogeneity across models and interaction histories \ab{\cite{yan2025sharechat,zhao2024wildchat,mireshghallah2024trustnobot}}. Yet similar outcomes across turns may arise from stable properties of a conversation or from responses to earlier scrutiny. Dynamic-panel research documents the bias of short fixed-effects panels \cite{nickell1981,dhaene2015}, making it important to distinguish recurrence between conversations from ordering within them.

\noindent\ab{\textbf{Research gap.} Controlled experiments establish effects of individual cues, and evidence audits assess generated citations; neither shows how multiple verification-related features appear together in deployed conversation records or how they relate to checking and challenging across tasks and turns.}

\section{Data and Methods}

\subsection{Corpora, analysis unit, and selection}

ShareChat contains publicly shared conversations from ChatGPT, Gemini, Grok, Perplexity, and Claude (April 2023--October 2025) \cite{yan2025sharechat}. We analyze English exchanges from the first four platforms; Claude's export is incompatible with our common feature extraction. WildChat-4.8M contains opt-in ChatGPT conversations \cite{zhao2024wildchat,ai2wildchat48m}. We retain U.S.-English interactions through July 31, 2025. Collection mechanisms, models, and preserved interface fields differ. Supplementary Section~A details sampling and conversation lengths.

Our \emph{unit of analysis} is an \emph{assistant--user response pair}: an assistant answer joined to the immediately following user message in the same conversation. Unanswered assistant turns are excluded. Of 489,308 candidate pairs, 489,283 receive valid outcome labels, spanning 98,007 conversations. A pair is \emph{qualifying} when a user replies to a checkable assistant claim or recommendation. Qualifying replies are labeled \emph{accepting}, \emph{checking}, or \emph{challenging}; all others are \emph{not qualifying}. We define \emph{scrutiny} as checking or challenging: requesting support for, or disputing, an answer. The scrutiny rate is their share of qualifying pairs; it does not measure successful verification.

There are 368,644 qualifying pairs (75.3\% of labeled pairs): 250,523 from ShareChat and 118,121 from WildChat (Table~\ref{tab:sample}). Behavioral analyses are conditional on an engaged qualifying exchange; they do not model initial engagement. Mean length counts pairs, not messages. Overall RQ1 rates and RQ2--RQ4 use qualifying pairs, while platform prevalence uses all labeled pairs. Source-preservation audits use platform-specific fields and denominators. Supplementary Sections~A and~C document these sample distinctions.

\begin{table}[t]
\centering\footnotesize
\caption{Behavioral analysis sample. Mean length counts observed assistant--user response pairs per conversation. Scrutiny is the percentage of qualifying pairs labeled checking or challenging. Platform-level means and medians appear in Supplementary Section~A.}
\label{tab:sample}
\setlength{\tabcolsep}{3pt}
\begin{tabular}{@{}lrrrrr@{}}
\toprule
Corpus & Pairs & Convs. & Mean & Qualifying & Scrutiny\\
\midrule
ShareChat & 311,591 & 44,847 & 6.9 & 250,523 & 30.3\%\\
WildChat & 177,692 & 53,160 & 3.3 & 118,121 & 25.2\%\\
\midrule
Total & 489,283 & 98,007 & 5.0 & 368,644 & 28.7\%\\
\bottomrule
\end{tabular}
\end{table}

\subsection{Operationalizing a response-level verification environment}

The \emph{verification environment} comprises evidence-related material, language, and possible user actions surrounding an assistant response. Its components have different functions. \emph{Material verification supports} are potential objects of inspection, such as explicit URLs, formatted hyperlinks, and source objects retained in platform metadata. A link permits potential inspection but does not establish claim-level support. \emph{Verification signifiers} are words that suggest evidentiary grounding or invoke authority, such as references to ``research'' or ``experts,'' even when no identifiable source is provided. Such language can appear without inspectable evidence.

\emph{Epistemic communication cues} express how the assistant presents its knowledge and limits. These include uncertainty, confidence, disclaimers, refusals, and exported process-like text; they do not establish correctness. \emph{Verification pathways} are opportunities for the user to seek more support or contest a response, such as following a link, asking the assistant to confirm a claim, or offering a correction. An assistant's verification prompt suggests a pathway; the next user message reveals conversational checking or challenging. The exports do not record source clicks or independent browsing. The first three describe responses; pathways describe possible and observed actions. We keep these roles separate, consistent with work on measurement validity \cite{jacobs2021measurement}, and do not combine them into a single transparency score. Supplementary Section~C details platform-specific source fields, and Section~E gives the complete feature definitions.

\subsection{Measuring the response and subsequent user action}\label{sec:measures}

\paragraph{Assistant-side indicators.}

We code eleven binary response features, recording their presence in text or preserved fields. Three measure potential evidence access: an \emph{explicit URL} (a written web address), a \emph{markdown link} (hyperlinked text encoded in the response), and \emph{inspectable evidence} (either kind of link or a preserved platform source object). Five capture epistemic communication: \emph{uncertainty cues} (hedges or limits on certainty), \emph{confidence cues} (strong assertions of certainty), \emph{disclaimers} (conditions or limitations on the response), \emph{refusals} (declining to answer a request), and \emph{reasoning traces} (process-like text retained in a model-specific thinking or analysis field). Exported reasoning text does not reveal internal computation. Two indicators capture signifiers: \emph{broad attribution language} invokes research, reports, experts, or another outside authority; an \emph{evidence signifier} presents an assertion as documented or supported without necessarily naming an inspectable source. The final feature, \emph{verification-prompt language}, marks assistant suggestions to check evidence or seek confirmation. These measures do not test factual accuracy or citation quality (Supplementary Section~E).

\ab{URLs, markdown links, source objects, and retained reasoning fields are identified structurally; the other features use pre-specified \emph{lexicons}, meaning sets of text patterns matched against responses. We developed the lexicons through literature-derived candidate patterns followed by inspection and revision of matches in both corpora, freezing them before outcome modeling. The development records do not specify the number of examples inspected per feature or the number of pilot judges; we therefore report only the documented final audit design (Supplementary Sections~E and~F). A separate, narrower \emph{citation-like attribution} detector is used only for the audit of source preservation; it is not an additional regression feature (Supplementary Sections~C and~E). The indicators overlap: any explicit URL also counts as inspectable evidence, so the high $\phi=0.957$ reflects a partly definitional subset relationship; 91.6\% of inspectable-evidence cases include a URL. A proposed broad evidentiary-language detector was removed because it nearly duplicated attribution ($\phi=0.995$). We fit one focal feature per model using common controls. These associations include naturally co-occurring cues (Supplementary Section~C).}

\paragraph{Next-turn outcomes.}

The immediately following user message receives one of four mutually exclusive labels. \emph{Checking} means requesting evidence, an explanation of the grounds for a claim, or confirmation of its accuracy without supplying a competing answer (for example, ``What is your source?'' or ``Are you sure?''). \emph{Challenging} means contesting the answer, asserting an error, or proposing a specific correction or alternative. \emph{Accepting} means proceeding with the answer without a visible check or challenge; it does not require an explicit statement of agreement. \emph{Not qualifying} includes exchanges without an engaged checkable claim, unrelated replies, or acknowledgments too ambiguous to interpret. For mixed acts, challenging takes precedence over checking, then accepting. \emph{Scrutiny} combines the checking and challenging labels. Labels identify visible acts, not private trust or successful verification. Observed checking is assistant-directed; external verification may appear as challenging or accepting, depending on the next reply. Supplementary Section~F provides decision boundaries and examples.

\paragraph{Automated classification and controls.}

We assign next-turn outcomes with Qwen3-30B-A3B, using the assistant answer, subsequent user message, and recent conversational context, under a frozen prompt, structured output schema, and deterministic decoding. The context limits, truncation flags, and recorded configuration are in Supplementary Section~A. We separately classify the conversation's \emph{first user message} for \emph{topic} (the subject of the initial request) and \emph{intent} (whether the request is Asking, Doing, or Expressing), using Qwen3-235B-A22B-Instruct-2507 and the taxonomy of Chatterji et al. \cite{chatterji2025how}. The seven topics are Writing, Practical Guidance, Seeking Information, Technical Help, Multimedia, Self-Expression, and Other/Unknown. These fixed opening-turn controls precede the outcome but can miss later topic drift. Further classifier details appear in Supplementary Sections~A and~F.

\paragraph{Manual validation.}

\ab{We assessed whether the classifier's assigned categories matched the codebook through a blinded positive-case audit by one human coder. Of 200 classifier-positive cases per qualifying outcome, estimated precision was 0.89 for checking and 0.92 for challenging (Supplementary Section~F). These figures measure agreement with one reviewer's decisions among model positives. They do not establish whether independent humans consistently separate the categories: no double-coded sample or human--human agreement estimate is available. Nor does positive-case sampling identify recall or the full confusion matrix. The codebook's competing-assertion rule makes the conceptual boundary explicit, while reproducibility of human application remains an open measurement-validity question.}

\subsection{Statistical design and estimation}\label{sec:statistical-design}

\ab{We address the four research questions with complementary analyses: descriptive prevalence and source-field audits (RQ1), pooled multinomial models of next-turn behavior (RQ2), task- and corpus-stratified models with held-out prediction (RQ3), and conversation-level clustering and sequence analyses (RQ4). The four RQs are introduced in Section~1; here we specify the estimators and comparison families. }

All behavioral models use qualifying assistant--user response pairs, with accepting as the reference category where relevant. We index pairs by $i$ and conversations by $c$, and distinguish this qualifying sample from the larger samples used for descriptive platform and source-field analyses (Section~3.1 and Supplementary Section~A).

\paragraph{RQ1: Feature prevalence, co-occurrence, and preserved sources.}

For each binary assistant-response feature $f$, let $X_{icf}=1$ when it is detected in pair $i$ of conversation $c$. We calculate its prevalence in the qualifying sample $\mathcal Q$ as

\begin{equation}\label{eq:rq1_prevalence}
\widehat p_f=\frac{1}{N_{\mathcal Q}}\sum_{(i,c)\in\mathcal Q}X_{icf},\qquad
\widehat p_{fs}=\frac{1}{N_s}\sum_{(i,c)\in\mathcal P_s}X_{icf},
\end{equation}

\ab{where $N_{\mathcal Q}=368{,}644$ and $\mathcal P_s$ contains labeled pairs from platform or corpus $s$. The platform-specific denominator $N_s$ therefore comes from the larger joined-pair sample, not the qualifying sample. We report feature counts alongside percentages to make these populations explicit. Binary-feature overlap is measured with the $\phi$ correlation, including the partly definitional URL/inspectable-evidence nesting (Supplementary Sections~C and~E).}

The Perplexity source-preservation audit uses assistant turns from the language-filtered ShareChat release, a different denominator from the paired behavioral sample. Let $A_i$ indicate narrow citation-like attribution, and let $B_i$ indicate an explicit URL, markdown link, or populated preserved source object (\texttt{source} or \texttt{source\_bar}). We estimate

\begin{equation}\label{eq:rq1_source_gap}
\widehat m_{\mathrm{Perplexity}}=
\frac{\sum_i A_i(1-B_i)}{\sum_i A_i},
\end{equation}

\ab{which is the share of citation-like turns without a source in the released record. For lexicon-detected attribution, we also calculate $\widehat p_{\mathrm{attr}}\times\widehat{\mathrm{precision}}$ under illustrative precision values. This sensitivity check estimates the share of \emph{detected} true positives; without recall it cannot recover the population prevalence of attribution. Source-field definitions, the overlap matrix, and the positive-case audit are in Supplementary Sections~C, E, and~F.}

\paragraph{RQ2: Pooled associations with checking and challenging.}

Let $Y_{ic}\in\{\mathrm{accept},\mathrm{check},\mathrm{challenge}\}$ denote the next-turn outcome. For each of the eleven retained response features, we fit a separate multinomial logistic model with accepting as the reference:

\begin{equation}\label{eq:multinom}
\log\!\left[\frac{\Pr(Y_{ic}=j)}{\Pr(Y_{ic}=\mathrm{accept})}\right]
=\alpha_j+\beta_{jf}X_{icf}+\boldsymbol\gamma_j^{\top}\mathbf Z_{ic},
\quad j\in\{\mathrm{check},\mathrm{challenge}\}.
\end{equation}

The common control vector $\mathbf Z_{ic}$ contains seven-category topic and three-category intent from the first user turn, turn-position bins (1, 2, 3, 4--5, 6--10, 11+), $\log(1+\text{assistant-response word count})$, model/system fixed effects (with sparse versions grouped), and indicators for truncated context, assistant answer, and user reply. Conversation-clustered covariance estimates account for repeated pairs from the same conversation. Estimating one focal feature at a time avoids simultaneous adjustment for indicators that substantially overlap; $\beta_{jf}$ describes the feature as it occurs with other cues in observed responses.

\ab{We report $\exp(\beta_{jf})$ as a relative risk ratio (RRR): the multiplicative change in $\Pr(Y=j)/\Pr(Y=\mathrm{accept})$ associated with the feature, holding the controls fixed. RRR $>1$ indicates a higher relative probability of $j$ versus acceptance. We report conversation-clustered 95\% confidence intervals. To establish whether a feature distinguishes checking from challenging, we directly test}

\begin{equation}\label{eq:rq2_contrast}
\begin{aligned}
\Delta_f&=\beta_{\mathrm{check},f}-\beta_{\mathrm{challenge},f},\\
\widehat{\operatorname{Var}}(\widehat\Delta_f)
&=\widehat{\operatorname{Var}}(\widehat\beta_{\mathrm{check},f})
+\widehat{\operatorname{Var}}(\widehat\beta_{\mathrm{challenge},f})
-2\widehat{\operatorname{Cov}}(\widehat\beta_{\mathrm{check},f},\widehat\beta_{\mathrm{challenge},f}).
\end{aligned}
\end{equation}

The Wald statistic $W_f=\widehat\Delta_f^{\,2}/\widehat{\operatorname{Var}}(\widehat\Delta_f)$ is compared with a $\chi^2_1$ distribution. The ratio $\exp(\Delta_f)=\mathrm{RRR}_{\mathrm{check},f}/\mathrm{RRR}_{\mathrm{challenge},f}$ exceeds one when the association leans toward checking and falls below one when it leans toward challenging. We apply Benjamini--Hochberg false-discovery-rate (FDR) correction to the nine stable pooled direct contrasts \cite{benjamini1995}. Sparse URL and inspectable-evidence cells produce unstable checking estimates under fine model-version fixed effects; these direct contrasts are marked non-estimable (NE). Supplementary Section~B reports a coarser platform-fixed-effect sensitivity analysis.

\paragraph{RQ3: Associations across tasks and corpora.}

\ab{We repeat the multinomial specification separately within seven opening-topic groups and the two corpora, using six prespecified features: URL, inspectable evidence, uncertainty, reasoning trace, disclaimer, and refusal. Within each stratum $g$, we refit Equation~\ref{eq:multinom} with stratum-specific coefficients and controls that vary within $g$ (for example, opening-topic indicators are omitted within a single topic). Conversation-clustered inference and the covariance-aware contrast remain the same:}
\begin{equation}\label{eq:rq3_strata}
\ab{\Delta_{fg}=\beta_{\mathrm{check},fg}-\beta_{\mathrm{challenge},fg}.}
\end{equation}
\ab{These are \emph{within-stratum} tests; significance in one stratum and non-significance in another do not constitute a test of an interaction between strata.}

\ab{The seven topic groups and two corpora yield $9\times6=54$ planned direct contrasts and $9\times6\times2=108$ component coefficients (each outcome versus acceptance). The six Multimedia direct contrasts are non-estimable; among the remaining 48, 32 meet the primary numerical-stability criterion, two yield finite but unstable test statistics, and 14 are non-estimable. We apply BH-FDR to the 108 planned component coefficients and, separately, to the 32 stable direct contrasts. Our main substantive claims use the more conservative 48-test non-Multimedia family, which retains the two finite-but-unstable $p$-values and assigns $p=1$ to 14 non-estimable cases; the 32-test correction is reported as a secondary sensitivity analysis. Supplementary Section~G provides the complete testing-family table, estimates, stability flags, and corrected values.}

\ab{For predictive comparison, five conversation-grouped folds evaluate six predictor blocks spanning response features, opening topic/intent/turn position, model/system, length, and truncation. We report macro one-versus-rest AUC, multiclass log loss $-N_{\mathrm{test}}^{-1}\sum_i\log\widehat{\Pr}(Y_i)$, and foldwise SDs. Supplementary Section~H specifies the blocks and full results.}

\paragraph{RQ4: Between-conversation clustering and sequential recurrence.}

For the $t$-th \emph{qualifying} pair in conversation $c$, define scrutiny as $S_{ct}=\mathbf{1}\{Y_{ct}\in\{\mathrm{check},\mathrm{challenge}\}\}$; accepting is coded zero. We first decompose scrutiny into conversation- and pair-level components,

\begin{equation}\label{eq:rq4_icc}
S_{ct}=\alpha+u_c+\varepsilon_{ct},\qquad
\mathrm{ICC}=\frac{\sigma_u^2}{\sigma_u^2+\sigma_\varepsilon^2},
\end{equation}

\ab{where $u_c$ is the conversation component. The intraclass correlation (ICC) measures the share of variation attributable to differences between conversations. The reported adjusted ICC applies covariate residualization using opening topic, intent, position, model/system, length, and truncation, followed by a between/within-conversation variance decomposition. Conversation identifiers are \emph{not} part of that covariate adjustment: removing conversation fixed effects would remove the between-conversation component the ICC is intended to describe.}

We next examine whether scrutiny recurs after a previous qualifying exchange. For pairs with a preceding qualifying pair in the same conversation, we estimate pooled and conversation-fixed-effect linear probability models:

\begin{equation}\label{eq:rq4_lag}
\begin{aligned}
S_{ct}&=\alpha+\rho S_{c,t-1}+\boldsymbol\gamma^{\top}\mathbf Z_{ct}+\varepsilon_{ct},\\
S_{ct}&=\alpha_c+\rho_{\mathrm{FE}}S_{c,t-1}
+\boldsymbol\gamma^{\top}\mathbf Z_{ct}+\varepsilon_{ct}.
\end{aligned}
\end{equation}

\ab{The lag coefficient is a conditional percentage-point association with scrutiny at the next qualifying exchange. The pooled model combines between- and within-conversation differences. In the fixed-effects specification, $\alpha_c$ absorbs opening topic and intent as well as any model/system identifier constant within that conversation; within-conversation predictors such as turn position, response length, and changing truncation flags remain identifiable. The lag advances between successive \emph{qualifying} pairs, even when nonqualifying exchanges occur between them. Short panels can bias dynamic fixed-effects estimates \cite{nickell1981}, so we compare samples with at least five, eight, and ten qualifying pairs. For the five-pair sample, a split-panel jackknife subtracts the mean half-panel FE estimate from twice the full-panel FE estimate \cite{dhaene2015}; conversation-bootstrap intervals describe uncertainty. Since short halves may be unstable, this is a sensitivity analysis (Supplementary Section~D).}

We separately study \emph{strictly adjacent} response pairs, now retaining all four labels, including not qualifying. For previous label $a$ and next label $b$, the empirical transition probability is

\begin{equation}\label{eq:rq4_transition}
\widehat T_{ab}=\frac{n(a\!\rightarrow\!b)}{\sum_{k}n(a\!\rightarrow\!k)}.
\end{equation}

\ab{Same-type excess repetition is measured as $E_{\mathrm{check}}=\widehat T_{\mathrm{check,check}}-\widehat T_{\mathrm{accept,check}}$ and $E_{\mathrm{challenge}}=\widehat T_{\mathrm{challenge,challenge}}-\widehat T_{\mathrm{accept,challenge}}$. Whole-conversation bootstrap resamples quantify uncertainty in their difference. Within-conversation shuffles preserve each conversation's four-outcome composition while randomizing order; comparing observed $E$ with the reported shuffle benchmark describes repetition beyond the conversation's outcome mix. The supplied summary records a benchmark, not the full permutation distribution, so we do not report a formal permutation $p$-value. We also compare the longest consecutive scrutiny run with its within-conversation permutation distribution.}

Finally, among conversations with at least five qualifying pairs, we calculate each conversation's scrutiny rate, $\overline S_c=T_c^{-1}\sum_{t=1}^{T_c}S_{ct}$, and designate the highest decile as \emph{scrutiny-heavy}. We report its unadjusted prevalence by opening topic, intent, and corpus and examine scrutiny across normalized conversation position. The strict-adjacency transitions and qualifying-pair lag models answer different sequence questions and use different denominators. Transition counts, bootstrap and permutation benchmarks, streaks, and position diagnostics appear in Supplementary Sections~D and~I.

\section{Results}\label{sec:results}
\subsection{\ab{RQ1: Grounding language outnumbers preserved inspectable support}\label{sec:rq1}}
\begin{figure}[!tbp]
\centering
\includegraphics[width=0.89\linewidth]{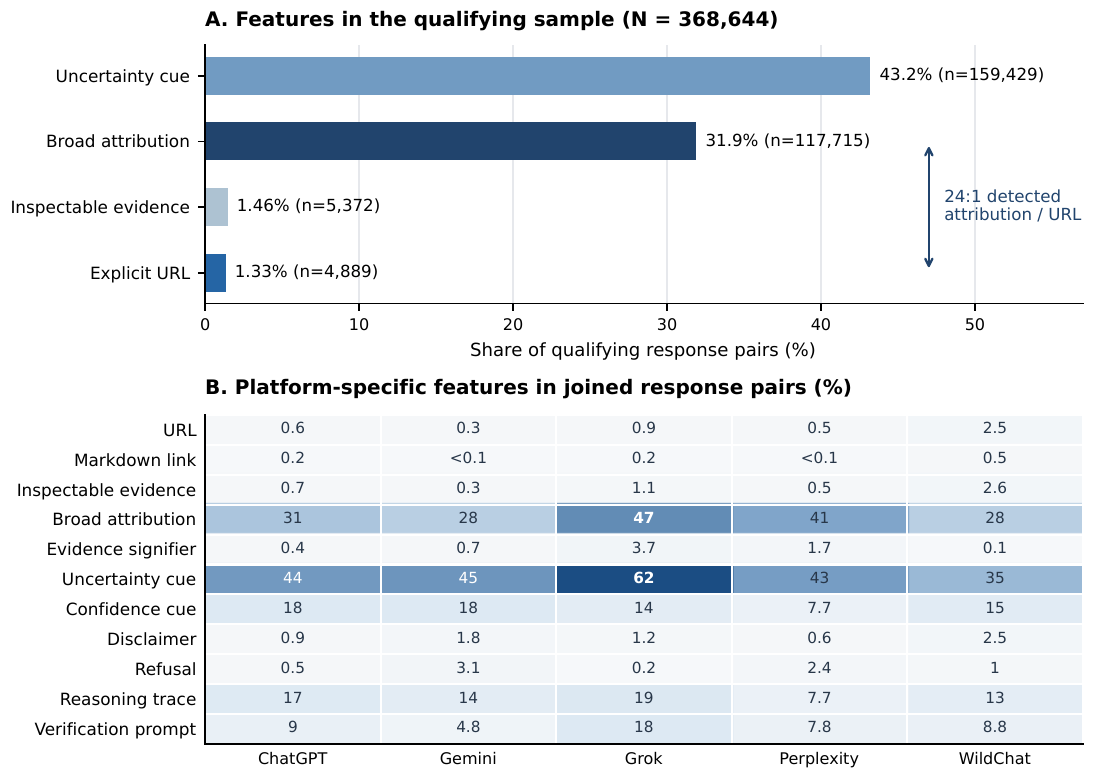}
\caption{\textbf{RQ1: Response features across samples and platforms.} (A) Detected feature prevalence among qualifying response pairs ($N=368{,}644$). The observed attribution-to-URL ratio is approximately 24:1; an illustrative sensitivity calculation accounting for attribution-detector precision yields approximately 10:1--20:1. (B) Platform-specific prevalence among all joined response pairs, with percentages shown in each cell. The denominators differ between panels. The measures describe features preserved in the released records and do not assess whether individual claims are adequately supported.}
\label{fig:rq1}
\end{figure}

\ab{In the released records, linguistic claims of evidentiary grounding are much more common than inspectable material. Among 368,644 qualifying response pairs, broad attribution is detected in 117,715 (31.9\%), compared with 4,889 explicit URLs (1.33\%) and 5,372 responses with any inspectable-evidence indicator (1.46\%; Figure~\ref{fig:rq1}A). The attribution-to-URL ratio is about 24:1. Broad attribution and explicit URLs are almost uncorrelated ($\phi=0.03$; Supplementary Section~C). Mentioning an authority and exposing a source are empirically distinct features in these records.}

\ab{The attribution detector's estimated precision is 0.67 in a small positive-case audit. Applied to detected positives, that estimate yields a share near 21.4\% and a ratio near 16:1 relative to URLs; the illustrative precision range 0.43--0.85 gives roughly 10:1--20:1. These are estimates of \emph{detected true positives} under assumed precision, with unknown recall (Supplementary Sections~E and~F).}

\ab{In Perplexity's available records, citation-like attribution occurs in 10.3\% of assistant turns, and 39.8\% of those turns have no preserved URL or source object (4.1\% of all turns). This quantifies missing source material in the released records (Supplementary Sections~A and~C). Platform profiles also differ: Grok shows 47\% attribution and 62\% uncertainty, while explicit URL rates are 0.5\% for Perplexity and 2.5\% for WildChat (Figure~\ref{fig:rq1}B; Supplementary Sections~C and~I).}

\noindent\ab{In the released records, language invoking authority is largely separate from material that can be inspected. A transparency measure that treats the two as equivalent would reward how convincingly an answer sounds sourced, even where little evidence is provided to examine.}

\subsection{\ab{RQ2: Response features accompany both checking and challenging}\label{sec:rq2}}
\begin{figure}[!tbp]
\centering
\includegraphics[width=0.92\linewidth]{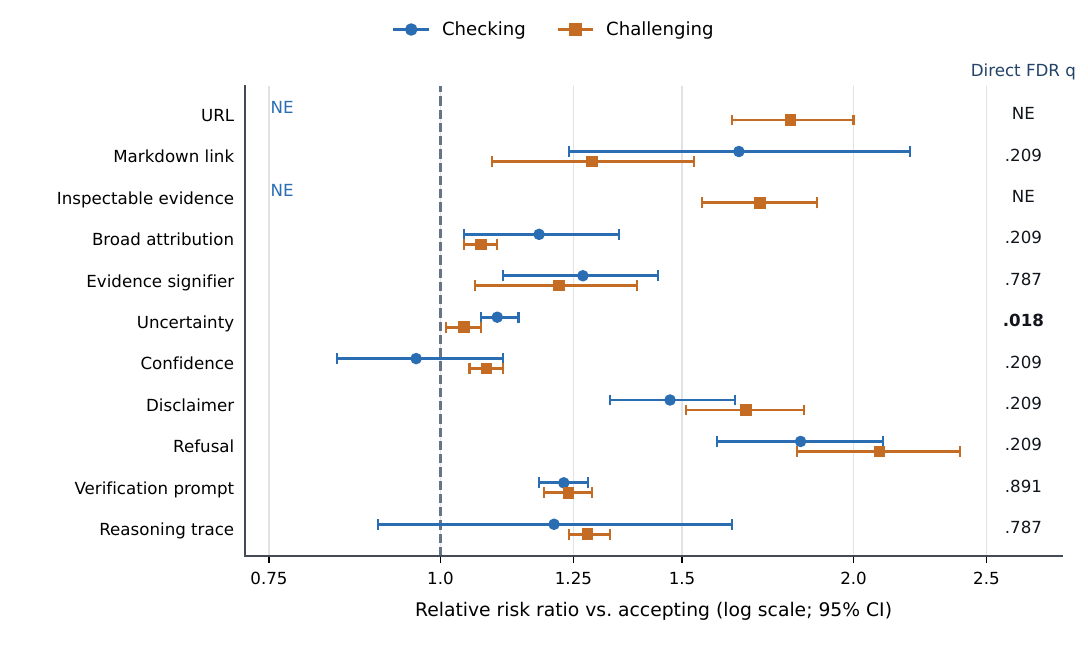}
\caption{\textbf{RQ2: Pooled associations between response features and subsequent user behavior.} Relative risk ratios for checking (blue) and challenging (orange), each compared with accepting, with conversation-clustered 95\% confidence intervals. All models use the common control set. NE indicates a coefficient that could not be stably estimated under the fine model-version specification. The right column reports FDR-adjusted $q$ values from covariance-aware direct checking-versus-challenging Wald tests. The direct contrast cannot be determined by visually comparing the separate confidence intervals.}
\label{fig:rq2}
\end{figure}

\ab{A request for evidence and a direct objection represent different next-turn actions. The pooled models nevertheless associate broad attribution, evidence signifiers, disclaimers, refusals, and verification prompts with both outcomes relative to acceptance (Figure~\ref{fig:rq2}). The distinction becomes visible only when checking and challenging are analyzed separately.}

\ab{Uncertainty is the only feature with a statistically distinguishable pooled checking-versus-challenging association after nine-test correction. Its RRRs relative to acceptance are 1.10 (95\% CI [1.07, 1.14]) for checking and 1.04 ([1.01, 1.07]) for challenging. The direct ratio is 1.06 ([1.02, 1.10], $q=0.018$): a 6\% ratio difference, modest in magnitude. Verification prompts show similar RRRs (1.23 and 1.24), while partial refusals accompany checking (1.83) and challenging (2.09).}

\ab{Platform-fixed-effect models show positive associations for URLs (1.73 checking; 1.86 challenging) and inspectable evidence (1.66; 1.77). Sparse features are less precise: markdown links yield a check/challenge ratio of 1.28 ([0.95, 1.73]; 1,133 positive pairs), and refusals 0.88 ([0.73, 1.04]; 2,159 pairs); verification prompts have a tighter interval (0.95--1.05). The intervals remain compatible with meaningful differences for sparse features (Supplementary Section~B, pooled estimates, feature counts, and platform-fixed-effect sensitivity).}

\noindent\ab{Checking and challenging are different responses to an answer, even when they follow similar cues. Expressions of uncertainty show a slight tendency toward checking, but the broader pattern is that the wording of an answer gives limited insight into whether a user will ask for support or dispute a claim. Keeping the two actions separate reveals that distinction.}

\subsection{\ab{RQ3: Scrutiny associations vary across tasks and corpora}\label{sec:rq3}}
\begin{figure}[!t]
\centering
\includegraphics[width=0.91\linewidth]{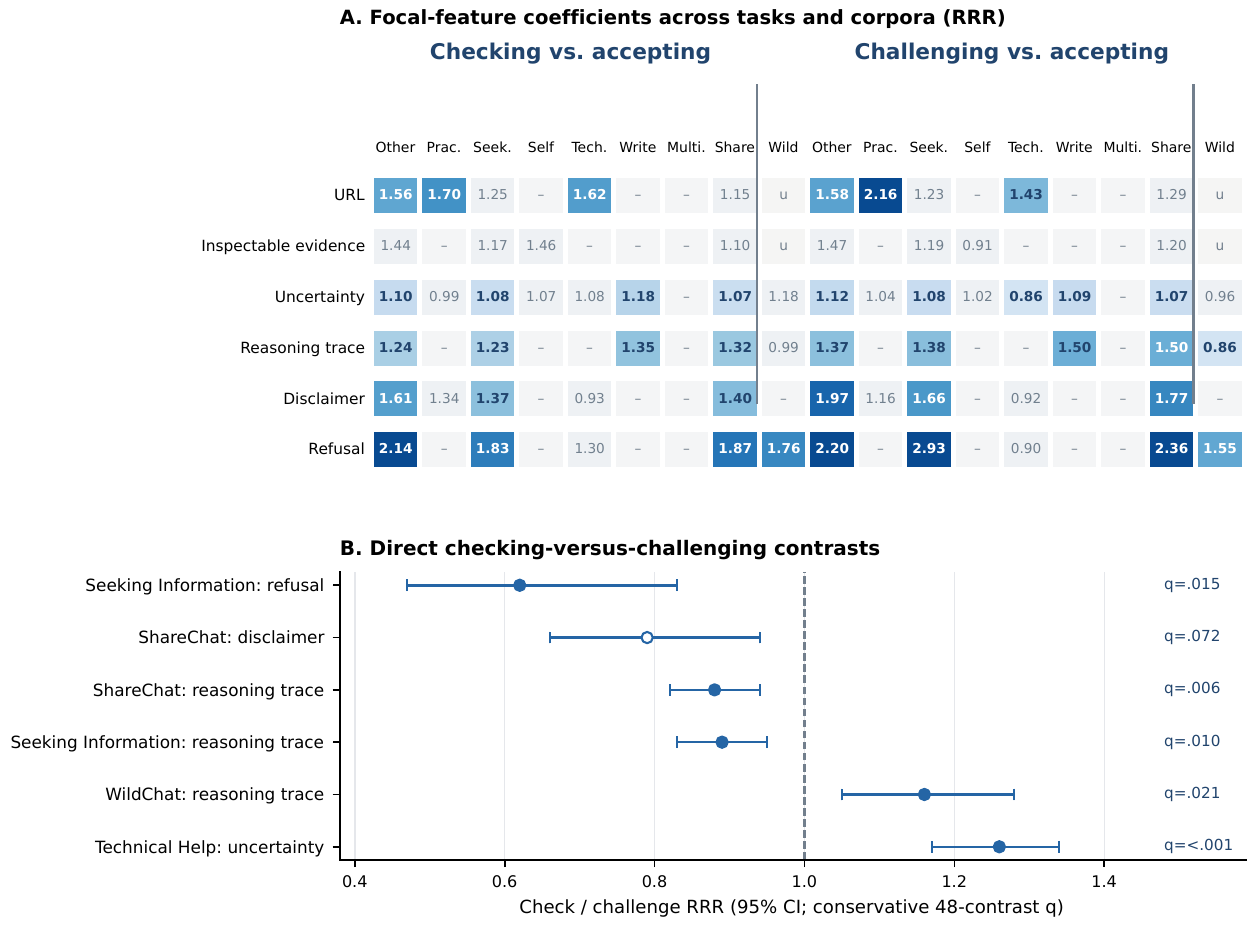}
\caption{\ab{\textbf{RQ3: Context-specific coefficients and direct contrasts.} (A) RRRs for checking and challenging versus accepting (108 planned tests). Columns: Other, Practical, Seeking, Self-Expression, Technical, Writing, Multimedia, ShareChat, WildChat. Bold blue: BH-FDR significant; gray: nonsignificant; dash: non-estimable; $u$: unstable. (B) Direct check/challenge ratios (95\% CIs); below one favors challenging, above one favors checking. Five filled points survive the conservative 48-test BH correction. Within-stratum significance does not test differences between strata.}}
\label{fig:rq3}
\end{figure}

\ab{The direct associations vary by task (Figure~\ref{fig:rq3}). In Technical Help, uncertainty has a checking/accepting RRR of 1.08 and a challenging/accepting RRR of 0.86; its direct check/challenge ratio is 1.26 (95\% CI [1.17, 1.34], conservative $q<0.001$). In Seeking Information, refusals accompany checking (1.83) and, more strongly, challenging (2.93), yielding a direct ratio of 0.62 ([0.47, 0.83], $q=0.015$). Reasoning traces in Seeking Information likewise lean toward challenging (0.89, [0.83, 0.95], $q=0.010$). Task context clarifies which form of scrutiny accompanies these cues.}

\ab{The reasoning-trace contrast reverses direction between collections: ShareChat favors challenging relative to checking (ratio 0.88, 95\% CI [0.82, 0.94], conservative $q=0.006$), whereas WildChat favors checking (1.16, [1.05, 1.28], $q=0.021$). Export formats, user populations, and model mix differ across the collections. Five direct contrasts pass the conservative 48-test correction. The ShareChat disclaimer contrast passes only the secondary 32-test family (ratio 0.79, $q=0.048$; conservative $q=0.072$). Full contrasts and component estimates appear in Supplementary Section~G.}

\ab{Task context also provides more predictive information than response features alone: conversation-grouped macro AUC is 0.592 (SD 0.005) for opening topic, intent, and position, versus 0.551 (0.004) for response features (Supplementary Section~H, predictive-comparison table).}

\Needspace{5\baselineskip}
\noindent\ab{The relationship between a response cue and the user's next action depends on the setting. Uncertainty accompanies checking more strongly in technical exchanges, while refusals and reasoning traces are associated with different forms of scrutiny in other contexts. The contrasting reasoning-trace patterns across the corpora also show why one overall estimate can conceal important differences in everyday use.}

\FloatBarrier
\subsection{\ab{RQ4: Conversation differences account for much of the observed recurrence of scrutiny}\label{sec:rq4}}
\ab{The cross-sectional contrasts leave another question: do skeptical replies repeat because of the order of turns, or because certain conversations contain more scrutiny overall? The sequence analyses separate these two patterns.}

\begin{figure}[!tbp]
\centering
\includegraphics[width=.74\linewidth]{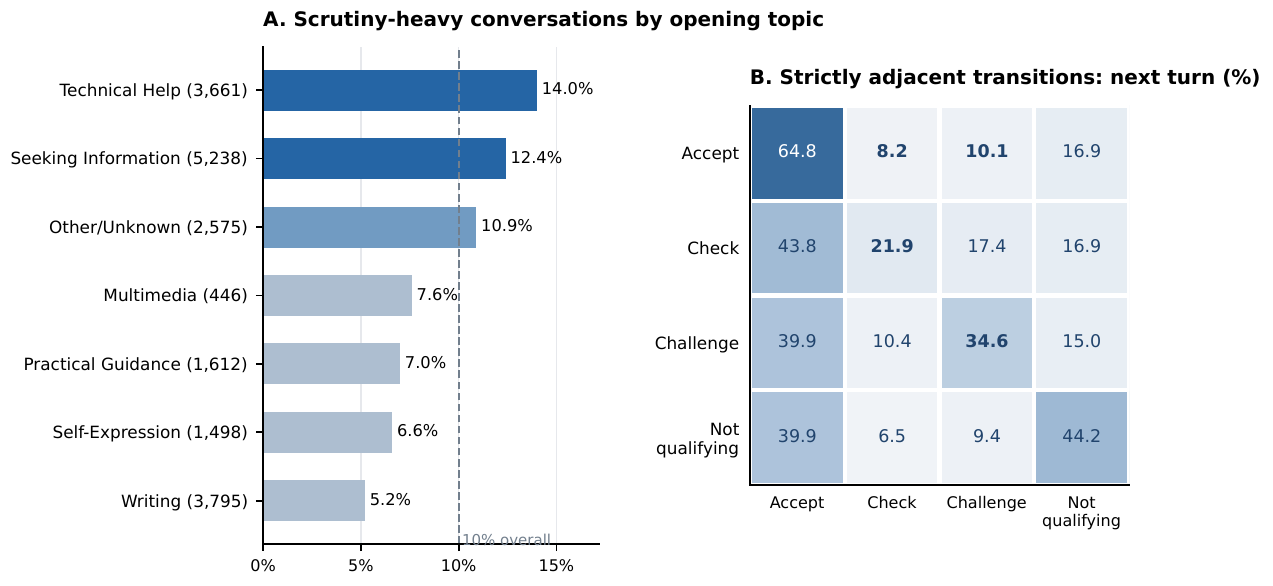}
\caption{\textbf{RQ4: Concentration of scrutiny and adjacent-turn behavior.} (A) Percentage of scrutiny-heavy conversations by opening topic among conversations with at least five qualifying exchanges. Parentheses show eligible conversation counts; the dashed 10\% line represents the top-decile benchmark by construction. Topic differences are unadjusted. (B) Row percentages for strictly adjacent next-turn outcomes, retaining accepting, checking, challenging, and not-qualifying labels.}
\label{fig:rq4-context}
\end{figure}

\ab{Checking and challenging are unevenly distributed. The overall scrutiny rate is 28.7\% of qualifying pairs, yet the highest decile among 18,847 conversations with at least five qualifying exchanges begins at a conversation-level rate of 70.63\% (1,885 conversations; Supplementary Section~D). The raw ICC is 0.207 (95\% CI [0.201, 0.214]), and the covariate-adjusted ICC is 0.184 ([0.178, 0.190]). High-scrutiny conversations are also more frequent in Technical Help (14.0\%) and Seeking Information (12.4\%) than Writing (5.2\%; Figure~\ref{fig:rq4-context}A). A pooled turn-level average therefore conceals pronounced variation across conversations.}

\begin{figure}[!tbp]
\centering
\includegraphics[width=.62\linewidth]{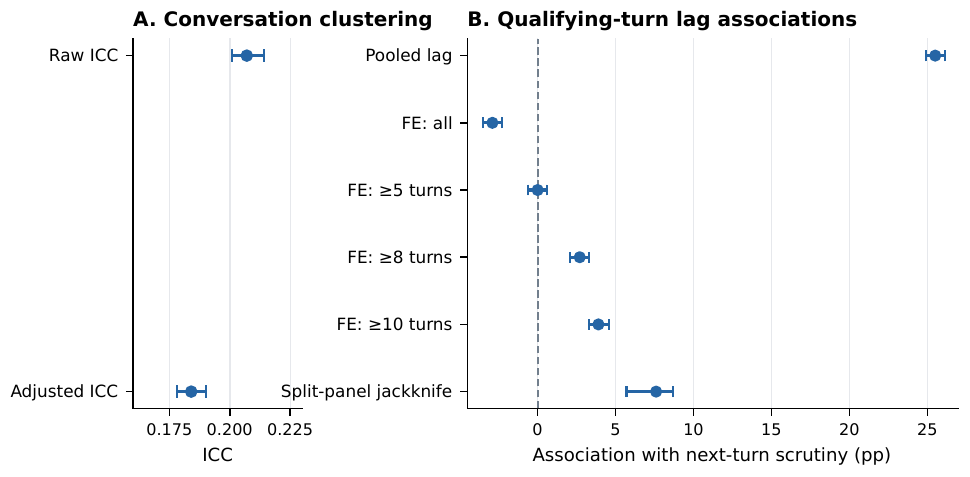}
\caption{\textbf{RQ4: Pooled and within-conversation sequential associations.} Lag estimates are shown as percentage-point associations between scrutiny at successive qualifying exchanges. Conversation fixed effects separate within-conversation associations from stable differences across conversations. Longer-panel restrictions and the split-panel jackknife assess sensitivity to dynamic-panel bias. The raw and adjusted ICCs are 0.207 and 0.184, respectively.}
\label{fig:rq4}
\end{figure}

\ab{Scrutiny at the preceding qualifying exchange has a +25.5-percentage-point association with scrutiny in the pooled lag model (95\% CI [24.9, 26.1]). The all-panel conversation-fixed-effect estimate is $-2.9$ points ([$-3.5$, $-2.3$]), and the estimate among conversations with at least five qualifying exchanges is near zero. Estimates become +2.7 ([2.1, 3.3]) at eight or more turns and +3.9 ([3.3, 4.6]) at ten or more; a split-panel jackknife in the five-turn sample gives +7.6 (bootstrap CI [5.7, 8.7]), subject to very short half-panels (Figure~\ref{fig:rq4}; Supplementary Section~D). The pooled estimate largely reflects differences across conversations.}

\ab{Strictly adjacent-turn transitions provide an ordering benchmark (Figure~\ref{fig:rq4-context}B). Checking repeats 13.6 percentage points more often after checking than after accepting, and challenging repeats 24.6 points more often after challenging than after accepting. Composition-preserving shuffles produce respective excesses of 10.8 and 18.3 points. The remaining order-associated excess is approximately 2.8 points for checking and 6.3 for challenging, about one-fifth and one-quarter of their observed excesses. Additional sequence diagnostics appear in Supplementary Sections~D and~I.}

\noindent\ab{Repeated checking and challenging are concentrated in particular conversations, while the sequence of turns adds a smaller share of the observed repetition. Looking only at individual turns conceals exchanges in which asking for evidence or disputing an answer becomes a recurring part of the interaction.}

\section{Discussion}

\subsection{\ab{Credibility signals and evidence accountability}}

\ab{The distinction between claiming evidentiary support and making evidence available matters for accountability. An answer that refers to an official report sounds grounded, but the user faces a different situation when the report is identified and can be examined. Our findings show how easily these two properties diverge in recorded LLM responses. A transparency measure that gives credit for either one rewards the appearance of support as readily as the availability of a source.}

\ab{For platform teams and independent auditors, the consequence is straightforward: claims of authority and preserved sources should be assessed separately. Otherwise, a system could improve its reported transparency simply by invoking research or experts more often. Researchers should also distinguish the presence of a source from the quality of the support it provides. Checking whether a cited document supports a particular claim is a valuable next layer of evidence auditing, built on a clear account of what material was available in the first place.}

\subsection{\ab{Checking and challenging reveal different verification needs}}

\ab{When users question an assistant, the kind of question matters. Asking for the basis of a claim puts the burden on the assistant to identify supporting evidence. Offering a correction puts the accuracy of the claim itself in dispute. In the observed conversations, similar response cues accompany both actions, yet checking and challenging show distinct patterns of repetition. Combining them into a single measure of skepticism loses information about what the user is asking the system to address.}

\ab{This distinction suggests a more useful way to assess follow-up behavior. Evidence requests call for traceable support connected to the claim. Challenges call for engagement with the proposed correction and a reconsideration of the answer. Product evaluations could assess these exchanges separately, examining whether the subsequent assistant response addresses the user's actual objection. Researchers can use the distinction to compare ways of supporting verification without treating a reduction in follow-up questions as sufficient evidence that a user's concern was resolved.}

\subsection{\ab{Task and corpus context shape the interpretation of cues}}

\ab{The significance of an epistemic cue depends partly on what the user is trying to accomplish. In technical exchanges, uncertainty often accompanies a request to check or clarify an answer. In information-seeking exchanges, refusals and reasoning text are more closely associated with challenges. The different reasoning-trace patterns across ShareChat and WildChat reinforce that cue--response relationships emerge within particular mixtures of tasks, systems, and recorded interface features.}

\ab{A cue that appears helpful in a controlled study of one task may have a different relationship with user behavior elsewhere. Evaluations of citations, uncertainty, or explanations should therefore examine the tasks in which those features appear and the actions users take afterward. For practitioners, the user's expressed request offers more specific guidance than a blanket assumption about the effect of a hedge or explanation. For researchers, reporting results by meaningful context makes the scope of a finding clearer and supports comparisons across datasets with different collection procedures.}

\subsection{\ab{Conversation-level scrutiny and disaggregated evaluation}}

\ab{An average scrutiny rate obscures how differently conversations unfold. Some exchanges involve repeated requests for evidence or corrections, while others move forward with little visible questioning. The sequence analyses show that much of this concentration is associated with differences between conversations. Turn ordering contributes additional repetition, but it is a smaller part of the pattern. Verification is therefore worth examining as a feature of an interaction, alongside the response to any individual answer.}

\ab{For researchers and platform evaluators, this points to a more informative unit of assessment. Conversation-level reports can show where requests for evidence or disputes recur, and whether these episodes are concentrated in a subset of interactions. An evaluation can then examine how the assistant responds to those repeated concerns, including whether it provides the requested source or engages a correction. This approach brings persistent verification burdens into view that are easily lost in averages over individual turns.}

\section{Limitations and Scope of Inference}
\ab{ShareChat consists of publicly shared conversations and WildChat of opt-in interactions; their populations, model mixes, and preserved fields differ. Platform comparisons therefore describe the released records. Behavioral estimates concern checkable assistant responses followed by an engaged user reply. The accepting label captures proceeding without visible scrutiny, including cases where users may have checked outside the conversation.}

\ab{Measurement validation currently includes one coder's blinded review of classifier-positive cases: precision is 0.89 for checking and 0.92 for challenging. Human agreement and recall remain unmeasured; the small lexicon audit limits attribution-rate precision. Topic labels reflect the opening turn. Sequential results distinguish conversation-level clustering, within-conversation lag associations, and order-preserving composition benchmarks; short-panel corrections merit cautious interpretation. }

\section{Conclusion}
\ab{Across two large collections of LLM conversations, language invoking authority is far more common than preserved inspectable evidence and is almost uncorrelated with explicit URLs. Checking and challenging have different observable patterns, while scrutiny varies markedly across conversations. These results establish a measurement distinction central to AI accountability: evidence accessibility, users' expressed scrutiny, and its distribution across an interaction should be evaluated on their own terms. A transparency score that treats an authority phrase as equivalent to an inspectable source conflates different properties of the information environment.}

\section*{Generative AI Usage Statement}
\ab{Generative AI tools were used to assist with manuscript writing and editing and with modifications to analysis code. All analyses were designed, conducted, checked, and interpreted by the authors, who take responsibility for the reported results and final manuscript.}
\section*{Ethical Considerations}
The data were released for research and are analyzed in aggregate. We do not attempt re-identification, build individual vulnerability profiles, or reproduce potentially identifying conversation text. Examples used in dissemination should be screened and paraphrased. More broadly, verification features can distribute work unevenly: an interface that requires constant checking burdens users, while one that only signals grounding may create false reassurance. These potential consequences require direct user-centered evaluation before design recommendations are treated as established remedies.

\printbibliography

@article{lee2004trust,
  author  = {Lee, John D. and See, Katrina A.},
  title   = {Trust in Automation: Designing for Appropriate Reliance},
  journal = {Human Factors},
  volume  = {46},
  number  = {1},
  pages   = {50--80},
  year    = {2004},
  doi     = {10.1518/hfes.46.1.50_30392}
}

@article{hoff2015trust,
  author  = {Hoff, Kevin A. and Bashir, Masooda},
  title   = {Trust in Automation: Integrating Empirical Evidence on Factors that Influence Trust},
  journal = {Human Factors},
  volume  = {57},
  number  = {3},
  pages   = {407--434},
  year    = {2015},
  doi     = {10.1177/0018720814547570}
}

@article{kohn2021measurement,
  author  = {Kohn, Spencer C. and de Visser, Ewart J. and Wiese, Eva and Lee, Yi-Ching and Shaw, Tyler H.},
  title   = {Measurement of Trust in Automation: A Narrative Review and Reference Guide},
  journal = {Frontiers in Psychology},
  volume  = {12},
  pages   = {604977},
  year    = {2021},
  doi     = {10.3389/fpsyg.2021.604977}
}

@inproceedings{guo2024decision,
  author    = {Guo, Ziyang and Wu, Yifan and Hartline, Jason and Hullman, Jessica},
  title     = {A Decision Theoretic Framework for Measuring AI Reliance},
  booktitle = {Proceedings of the 2024 ACM Conference on Fairness, Accountability, and Transparency},
  year      = {2024},
  doi       = {10.1145/3630106.3658901}
}

@inproceedings{schemmer2023appropriate,
  author    = {Schemmer, Max and K{\"u}hl, Niklas and Benz, Carina and Bartos, Andrea and Satzger, Gerhard},
  title     = {Appropriate Reliance on {AI} Advice: Conceptualization and the Effect of Explanations},
  booktitle = {Proceedings of the 28th International Conference on Intelligent User Interfaces},
  pages     = {410--422},
  year      = {2023},
  doi       = {10.1145/3581641.3584066}
}

@inproceedings{kim2024notsure,
  author    = {Kim, Sunnie S. Y. and Liao, Q. Vera and Vorvoreanu, Mihaela and Ballard, Stephanie and Vaughan, Jennifer Wortman},
  title     = {``I'm Not Sure, But...'': Examining the Impact of Large Language Models' Uncertainty Expression on User Reliance and Trust},
  booktitle = {Proceedings of the 2024 ACM Conference on Fairness, Accountability, and Transparency},
  pages     = {822--835},
  year      = {2024},
  doi       = {10.1145/3630106.3658941}
}

@inproceedings{kim2025fostering,
  author    = {Kim, Sunnie S. Y. and Vaughan, Jennifer Wortman and Liao, Q. Vera and Lombrozo, Tania and Russakovsky, Olga},
  title     = {Fostering Appropriate Reliance on Large Language Models: The Role of Explanations, Sources, and Inconsistencies},
  booktitle = {Proceedings of the 2025 CHI Conference on Human Factors in Computing Systems},
  year      = {2025},
  doi       = {10.1145/3706598.3714020}
}

@article{vasconcelos2023explanations,
  author  = {Vasconcelos, Helena and J{\"o}rke, Matthew and Grunde-McLaughlin, Madeleine and Gerstenberg, Tobias and Bernstein, Michael S. and Krishna, Ranjay},
  title   = {Explanations Can Reduce Overreliance on AI Systems During Decision-Making},
  journal = {Proceedings of the ACM on Human-Computer Interaction},
  volume  = {7},
  number  = {CSCW1},
  pages   = {1--38},
  year    = {2023},
  doi     = {10.1145/3579605}
}

@inproceedings{ding2025citations,
  author    = {Ding, Yifan and Facciani, Matthew and Joyce, Ellen and Poudel, Amrit and Bhattacharya, Sanmitra and Veeramani, Balaji and Agui{\~n}aga, Sal and Weninger, Tim},
  title     = {Citations and Trust in LLM Generated Responses},
  booktitle = {Proceedings of the AAAI Conference on Artificial Intelligence},
  volume    = {39},
  number    = {22},
  pages     = {23787--23795},
  year      = {2025},
  doi       = {10.1609/aaai.v39i22.34550}
}

@article{bucinca2021trust,
  author  = {Bu{\c{c}}inca, Zana and Malaya, Maja Barbara and Gajos, Krzysztof Z.},
  title   = {To Trust or to Think: Cognitive Forcing Functions Can Reduce Overreliance on AI in AI-Assisted Decision-Making},
  journal = {Proceedings of the ACM on Human-Computer Interaction},
  volume  = {5},
  number  = {CSCW1},
  pages   = {1--21},
  year    = {2021},
  doi     = {10.1145/3449287},
  note    = {Article 188}
}

@book{gibson1979,
  author    = {Gibson, James J.},
  title     = {The Ecological Approach to Visual Perception},
  publisher = {Houghton Mifflin},
  year      = {1979}
}

@book{norman2013,
  author    = {Norman, Donald A.},
  title     = {The Design of Everyday Things: Revised and Expanded Edition},
  publisher = {Basic Books},
  year      = {2013}
}

@article{hutchby2001,
  author  = {Hutchby, Ian},
  title   = {Technologies, Texts and Affordances},
  journal = {Sociology},
  volume  = {35},
  number  = {2},
  pages   = {441--456},
  year    = {2001},
  doi     = {10.1177/S0038038501000219}
}

@article{rieh2002,
  author  = {Rieh, Soo Young},
  title   = {Judgment of Information Quality and Cognitive Authority in the Web},
  journal = {Journal of the American Society for Information Science and Technology},
  volume  = {53},
  number  = {2},
  pages   = {145--161},
  year    = {2002},
  doi     = {10.1002/asi.10017}
}

@article{metzger2007,
  author  = {Metzger, Miriam J.},
  title   = {Making Sense of Credibility on the Web: Models for Evaluating Online Information and Recommendations for Future Research},
  journal = {Journal of the American Society for Information Science and Technology},
  volume  = {58},
  number  = {13},
  pages   = {2078--2091},
  year    = {2007},
  doi     = {10.1002/asi.20672}
}

@article{hilligoss2008,
  author  = {Hilligoss, Brian and Rieh, Soo Young},
  title   = {Developing a Unifying Framework of Credibility Assessment: Construct, Heuristics, and Interaction in Context},
  journal = {Information Processing \& Management},
  volume  = {44},
  number  = {4},
  pages   = {1467--1484},
  year    = {2008},
  doi     = {10.1016/j.ipm.2007.10.001}
}

@article{pirolli1999information,
  author  = {Pirolli, Peter and Card, Stuart},
  title   = {Information Foraging},
  journal = {Psychological Review},
  volume  = {106},
  number  = {4},
  pages   = {643--675},
  year    = {1999},
  doi     = {10.1037/0033-295X.106.4.643}
}

@article{sperber2010,
  author  = {Sperber, Dan and Cl{\'e}ment, Fabrice and Heintz, Christophe and Mascaro, Olivier and Mercier, Hugo and Origgi, Gloria and Wilson, Deirdre},
  title   = {Epistemic Vigilance},
  journal = {Mind \& Language},
  volume  = {25},
  number  = {4},
  pages   = {359--393},
  year    = {2010},
  doi     = {10.1111/j.1468-0017.2010.01394.x}
}

@article{liao2021question,
  author  = {Liao, Q. Vera and Pribi{\'c}, Milo{\v{s}} and Han, Jaesik and Miller, Sarah and Sow, Daby},
  title   = {Question-Driven Design Process for Explainable AI User Experiences},
  journal = {arXiv preprint arXiv:2104.03483},
  year    = {2021}
}

@article{liao2024transparency,
  author  = {Liao, Q. Vera and Vaughan, Jennifer Wortman},
  title   = {AI Transparency in the Age of LLMs: A Human-Centered Research Roadmap},
  journal = {Harvard Data Science Review},
  note    = {Special Issue 5},
  year    = {2024},
  doi     = {10.1162/99608f92.8036d03b}
}

@techreport{chatterji2025how,
  author      = {Chatterji, Aaron and Cunningham, Tom and Deming, David J. and Hitzig, Zo{\"e} and Ong, Christopher and Shan, Carl Y. and Wadman, Kevin},
  title       = {How People Use ChatGPT},
  institution = {National Bureau of Economic Research},
  type        = {Working Paper},
  number      = {34255},
  year        = {2025},
  doi         = {10.3386/w34255}
}

@inproceedings{zhao2024wildchat,
  author    = {Zhao, Wenting and Ren, Xiang and Hessel, Jack and Cardie, Claire and Choi, Yejin and Deng, Yuntian},
  title     = {WildChat: 1M ChatGPT Interaction Logs in the Wild},
  booktitle = {International Conference on Learning Representations},
  year      = {2024}
}

@misc{ai2wildchat48m,
  author       = {{Allen Institute for AI}},
  title        = {WildChat-4.8M Dataset Card},
  howpublished = {Hugging Face Datasets},
  note         = {2025 dataset card; accessed August 5, 2026}
}

@article{yan2025sharechat,
  author  = {Yan, Yueru and Nguyen, Tuc and Su, Bo and Lieffers, Melissa and Le, Thai},
  title   = {ShareChat: A Dataset of Chatbot Conversations in the Wild},
  journal = {arXiv preprint arXiv:2512.17843},
  year    = {2025}
}

@article{benjamini1995,
  author = {Benjamini, Yoav and Hochberg, Yosef},
  title = {Controlling the False Discovery Rate: A Practical and Powerful Approach to Multiple Testing},
  journal = {Journal of the Royal Statistical Society: Series B (Methodological)},
  year = {1995},
  volume = {57},
  number = {1},
  pages = {289--300},
  doi = {10.1111/j.2517-6161.1995.tb02031.x}
}

@article{nickell1981,
  author = {Nickell, Stephen},
  title = {Biases in Dynamic Models with Fixed Effects},
  journal = {Econometrica},
  year = {1981},
  volume = {49},
  number = {6},
  pages = {1417--1426},
  doi = {10.2307/1911408}
}

@article{dhaene2015,
  author = {Dhaene, Geert and Jochmans, Koen},
  title = {Split-panel Jackknife Estimation of Fixed-effect Models},
  journal = {The Review of Economic Studies},
  year = {2015},
  volume = {82},
  number = {3},
  pages = {991--1030},
  doi = {10.1093/restud/rdv007}
}

@article{liao2026chatbutverify,
  author  = {Liao, Mengqi and Sundar, S. Shyam},
  title   = {Chat but Verify: Combating Misinformation in Conversational Generative {AI} with Verification Affordance},
  journal = {Journal of Computer-Mediated Communication},
  volume  = {31},
  number  = {3},
  pages   = {zmag012},
  year    = {2026},
  doi     = {10.1093/jcmc/zmag012}
}

@inproceedings{mayerhofer2025blending,
  author    = {Mayerhofer, Kerstin and Capra, Rob and Elsweiler, David},
  title     = {Blending Queries and Conversations: Understanding Trust, Verification, and System Choice in Search and Chat Interactions},
  booktitle = {Proceedings of the 2025 ACM SIGIR Conference on Human Information Interaction and Retrieval (CHIIR)},
  year      = {2025},
  pages     = {168--178},
  publisher = {ACM},
  doi       = {10.1145/3698204.3716454}
}

@inproceedings{mireshghallah2024trustnobot,
  author    = {Mireshghallah, Niloofar and Antoniak, Maria and More, Yash and Choi, Yejin and Farnadi, Golnoosh},
  title     = {Trust No Bot: Discovering Personal Disclosures in Human-{LLM} Conversations in the Wild},
  booktitle = {Proceedings of the Conference on Language Modeling (COLM)},
  year      = {2024},
  note      = {arXiv:2407.11438}
}

@inproceedings{bansal2021whole,
  author    = {Bansal, Gagan and Wu, Tongshuang and Zhou, Joyce and Fok, Raymond and Nushi, Besmira and Kamar, Ece and Ribeiro, Marco Tulio and Weld, Daniel},
  title     = {Does the Whole Exceed Its Parts? The Effect of {AI} Explanations on Complementary Team Performance},
  booktitle = {Proceedings of the 2021 CHI Conference on Human Factors in Computing Systems},
  year      = {2021},
  doi       = {10.1145/3411764.3445717}
}

@incollection{sundar2008main,
  author    = {Sundar, S. Shyam},
  title     = {The {MAIN} Model: A Heuristic Approach to Understanding Technology Effects on Credibility},
  booktitle = {Digital Media, Youth, and Credibility},
  editor    = {Metzger, Miriam J. and Flanagin, Andrew J.},
  publisher = {MIT Press},
  pages     = {73--100},
  year      = {2008}
}

@article{metzger2013credibility,
  author  = {Metzger, Miriam J. and Flanagin, Andrew J.},
  title   = {Credibility and Trust of Information in Online Environments: The Use of Cognitive Heuristics},
  journal = {Journal of Pragmatics},
  volume  = {59},
  pages   = {210--220},
  year    = {2013},
  doi     = {10.1016/j.pragma.2013.07.012}
}

@article{wineburg2019lateral,
  author  = {Wineburg, Sam and McGrew, Sarah},
  title   = {Lateral Reading and the Nature of Expertise: Reading Less and Learning More When Evaluating Digital Information},
  journal = {Teachers College Record},
  volume  = {121},
  number  = {11},
  year    = {2019},
  doi     = {10.1177/016146811912101102}
}

@inproceedings{tankelevitch2024metacognitive,
  author    = {Tankelevitch, Lev and Kewenig, Viktor and Simkute, Auste and Scott, Ava Elizabeth and Sarkar, Advait and Sellen, Abigail and Rintel, Sean},
  title     = {The Metacognitive Demands and Opportunities of Generative {AI}},
  booktitle = {Proceedings of the 2024 CHI Conference on Human Factors in Computing Systems},
  year      = {2024},
  doi       = {10.1145/3613904.3642902}
}

@article{mielke2022linguistic,
  author  = {Mielke, Sabrina J. and Szlam, Arthur and Dinan, Emily and Boureau, Y-Lan},
  title   = {Reducing Conversational Agents' Overconfidence Through Linguistic Calibration},
  journal = {Transactions of the Association for Computational Linguistics},
  volume  = {10},
  pages   = {857--872},
  year    = {2022},
  doi     = {10.1162/tacl_a_00494}
}

@book{hyland1998hedging,
  author    = {Hyland, Ken},
  title     = {Hedging in Scientific Research Articles},
  publisher = {John Benjamins},
  address   = {Amsterdam},
  year      = {1998}
}

@inproceedings{liu2023verifiability,
  author    = {Liu, Nelson F. and Zhang, Tianyi and Liang, Percy},
  title     = {Evaluating Verifiability in Generative Search Engines},
  booktitle = {Findings of the Association for Computational Linguistics: EMNLP 2023},
  pages     = {7001--7025},
  year      = {2023},
  doi       = {10.18653/v1/2023.findings-emnlp.467}
}

@article{rashkin2023attribution,
  author  = {Rashkin, Hannah and Nikolaev, Vitaly and Lamm, Matthew and Aroyo, Lora and Collins, Michael and Das, Dipanjan and Petrov, Slav and Tomar, Gaurav Singh and Turc, Iulia and Reitter, David},
  title   = {Measuring Attribution in Natural Language Generation Models},
  journal = {Computational Linguistics},
  volume  = {49},
  number  = {4},
  pages   = {777--840},
  year    = {2023},
  doi     = {10.1162/coli_a_00486}
}

@inproceedings{gao2023alce,
  author    = {Gao, Tianyu and Yen, Howard and Yu, Jiatong and Chen, Danqi},
  title     = {Enabling Large Language Models to Generate Text with Citations},
  booktitle = {Proceedings of the 2023 Conference on Empirical Methods in Natural Language Processing},
  pages     = {6465--6488},
  year      = {2023},
  doi       = {10.18653/v1/2023.emnlp-main.398}
}

@inproceedings{venkit2024answerengines,
  author    = {Venkit, Pranav Narayanan and Laban, Philippe and Zhou, Yilun and Mao, Yixin and Wu, Chien-Sheng},
  title     = {Search Engines in the AI Era: A Qualitative Understanding to the False Promise of Factual and Verifiable Source-Cited Responses in LLM-based Search},
  booktitle = {Proceedings of the 2025 ACM Conference on Fairness, Accountability, and Transparency (FAccT '25)},
  pages     = {1325--1340},
  year      = {2025},
  doi       = {10.1145/3715275.3732089},
  publisher = {Association for Computing Machinery}
}

@inproceedings{jacobs2021measurement,
  title={Measurement and Fairness},
  author={Jacobs, Abigail Z. and Wallach, Hanna},
  booktitle={Proceedings of the 2021 ACM Conference on Fairness, Accountability, and Transparency},
  pages={375--385}, year={2021}, publisher={ACM},
  doi={10.1145/3442188.3445901}
}

@misc{mata2023sanctions,
  author = {{United States District Court for the Southern District of New York}},
  title = {{Mata v. Avianca, Inc.: Opinion and Order on Sanctions}},
  year = {2023},
  date = {2023-06-22},
  url = {https://law.justia.com/cases/federal/district-courts/new-york/nysdce/1%3A2022cv01461/575368/54/},
  note = {No. 22-cv-1461 (PKC), Document 54}
}
\appendix

\section{Corpus construction, sample flow, and conversation lengths}\label{app:data}
\subsection{Source releases and analytic restrictions}
The analysis combines two releases with different inclusion mechanisms. ShareChat consists of publicly shared conversations collected across ChatGPT, Gemini, Grok, Perplexity, and Claude from April 2023 to October 2025 \cite{yan2025sharechat}. The language-filtered ShareChat release used for the platform-schema inspection was downloaded on August 1, 2026 and contains 129,584 conversations: 94,086 from ChatGPT, 14,612 from Perplexity, 13,149 from Grok, 6,826 from Gemini, and 911 from Claude. The shared response-pair analysis uses the first four platforms, retains English-language exchanges with an assistant response followed immediately by a user message, and excludes Claude from the common extraction because of schema incompatibility. These full-release conversation counts are \emph{not} the denominators of the response-pair or behavioral models.

WildChat-4.8M consists of opt-in ChatGPT interactions \cite{zhao2024wildchat,ai2wildchat48m}. We use its public release through July 31, 2025, restricting conversations to U.S.-English interactions with the required assistant-to-user adjacency. The geography restriction applies only to WildChat. Corpus contrasts consequently combine differences in collection mechanisms, geography, model versions, and the preserved response representation.

\subsection{Response-pair construction and qualifying gate}
Our analytic unit is an assistant response joined to the \emph{immediately next user message}. Conversation length in the following tables counts these observed pairs, not every user or assistant message. For example, an assistant response without a subsequent user message cannot form a pair. There are 489,308 candidate pairs; 25 failed to receive a valid structured outcome label, leaving 489,283 labeled pairs from 98,007 conversations. Requiring a checkable claim or recommendation and a reply that engages it yields 368,644 qualifying pairs. A reply may be accepting, checking, or challenging only within that qualifying sample; otherwise it is labeled not qualifying. Scrutiny is checking or challenging among qualifying pairs.

\begin{table}[htbp]
\centering\small
\caption{Construction of the response-pair sample. Qualifying and scrutiny percentages use corpus-specific labeled pairs and qualifying pairs, respectively. Means divide labeled pairs by conversations with an observed pair; they do not represent total messages per conversation.}
\label{tab:supp_data_flow}
\setlength{\tabcolsep}{4pt}
\begin{tabular}{lrrrrrr}
\toprule
Corpus & Labeled pairs & Conversations & Mean pairs/conv. & Qualifying & Qualifying (\%) & Scrutiny (\%)\\
\midrule
ShareChat & 311,591 & 44,847 & 6.9 & 250,523 & 80.4 & 30.3\\
WildChat & 177,692 & 53,160 & 3.3 & 118,121 & 66.5 & 25.2\\
\midrule
Total & 489,283 & 98,007 & 5.0 & 368,644 & 75.3 & 28.7\\
\bottomrule
\end{tabular}
\end{table}

\paragraph{How to read the qualifying and scrutiny rates.}
The qualifying rate uses all successfully labeled response pairs as its denominator. The scrutiny rate uses only qualifying pairs: its numerator is the count of checking plus challenging labels, and its denominator is the count of accepting, checking, and challenging labels combined. Neither rate measures externally completed fact-checking. A qualifying reply can proceed with an answer (accepting) even if the user privately verified it; the transcript cannot distinguish that case from proceeding without external checking.

\subsection{Conversation-length distribution across platforms}
Conversation length is heterogeneous even within ShareChat. Table~\ref{tab:supp_lengths} reproduces the platform-level means and medians summarized in the proposal-defense sample-comparison chart. A mean is calculated over analyzed conversations belonging to the named platform, using observed response pairs; medians count the same unit. The medians show why a single overall average can obscure differences across products: Perplexity has a median of one observed pair, while the ShareChat ChatGPT, Gemini, and Grok groups each have a median of three. The exported summary does not provide platform-specific length quantiles, so we do not infer longer-tail percentiles. The original multi-panel sample comparison is retained in Figure~\ref{fig:new_sample}.

\begin{table}[htbp]
\centering\small
\caption{Observed assistant--user response pairs per conversation by platform, in the analyzed paired data. The platform means are rounded to one decimal and medians to whole pairs, as reported in the defense summary. These numbers are not derived from the full ShareChat release counts.}
\label{tab:supp_lengths}
\begin{tabular}{llrr}
\toprule
Corpus & Platform & Mean & Median\\
\midrule
ShareChat & ChatGPT & 7.5 & 3\\
ShareChat & Gemini & 11.2 & 3\\
ShareChat & Grok & 6.2 & 3\\
ShareChat & Perplexity & 1.6 & 1\\
WildChat & ChatGPT interactions & 3.3 & 2\\
\bottomrule
\end{tabular}
\end{table}

\subsection{Labeling inputs, time ordering, and scope of the source-field audit}
The outcome labeler uses Qwen3-30B-A3B with a fixed codebook and structured JSON output. The proposal's archived configuration specifies local Ollama serving of a Q4\_K\_M-quantized checkpoint, thinking mode disabled, temperature zero, top-$p=0.1$, an 8,192-token context window, and a 160-token maximum generation; the deployment record did not preserve the Ollama build identifier or specify a random seed. Each labeling call receives recent conversational context (capped at 2,500 characters), the focal assistant answer (5,000 characters), and the following user reply (4,000 characters). Context, assistant-answer, and user-reply truncation are recorded separately. The first user message of each conversation is labeled for topic and intent with Qwen3-235B-A22B-Instruct-2507, and those labels are carried forward unchanged to that conversation's later pairs. Topic and intent thus describe the opening task and cannot absorb information from a later outcome, but can miss topic changes across turns.

The platform-specific source inspection uses the \emph{full language-filtered ShareChat release}, and describes preserved export fields; the response-feature and outcome analyses use the smaller paired sample. Grok exports \texttt{links} and \texttt{source}, while Perplexity exports \texttt{source} and \texttt{source\_bar}. ChatGPT and Gemini do not provide directly analogous export fields in the inspected release. Source-field availability is not equivalent to a source being present on every turn or displayed in the live interface. Further source-field and feature-overlap diagnostics appear in Supplementary Section~C.

\section{Pooled outcomes and sensitivity analyses}\label{app:pooled}
\ab{Table~\ref{tab:pooled} reports the full pooled checking-versus-accepting and challenging-versus-accepting estimates and direct Wald contrasts, formerly shown in the main paper. Main Figure~2 displays the same component associations and corrected direct contrasts. Non-estimable (NE) primary link coefficients were not used in direct contrasts. The platform-fixed-effect sensitivity below reports the link-based coefficients under coarser controls.}

\begin{table}[htbp]
\centering\scriptsize
\caption{\ab{Pooled multinomial models on the qualifying sample (moved from the main paper). Entries are relative risk ratios (RRRs) with 95\% confidence intervals. The $q$ column reports BH-FDR correction across the nine stable direct checking-versus-challenging contrasts.}}
\label{tab:pooled}
\setlength{\tabcolsep}{3pt}
\begin{tabular}{lrrrr}
\toprule
\ab{Feature }& \ab{Check/accept }& \ab{Challenge/accept }& \ab{Check/challenge }& \ab{$q$}\\
\midrule
\ab{URL }& \ab{NE }& \ab{1.80 [1.63, 2.00] }& \ab{NE }& \ab{--}\\
\ab{Markdown link }& \ab{1.65 [1.24, 2.20] }& \ab{1.29 [1.09, 1.53] }& \ab{1.28 [0.95, 1.73] }& \ab{0.209}\\
\ab{Inspectable evidence }& \ab{NE }& \ab{1.71 [1.55, 1.88] }& \ab{NE }& \ab{--}\\
\ab{Broad attribution }& \ab{1.18 [1.04, 1.35] }& \ab{1.07 [1.04, 1.10] }& \ab{1.11 [0.97, 1.26] }& \ab{0.209}\\
\ab{Evidence signifier }& \ab{1.27 [1.11, 1.44] }& \ab{1.22 [1.06, 1.39] }& \ab{1.04 [0.88, 1.23] }& \ab{0.787}\\
\ab{Uncertainty }& \ab{1.10 [1.07, 1.14] }& \ab{1.04 [1.01, 1.07] }& \ab{1.06 [1.02, 1.10] }& \ab{0.018}\\
\ab{Confidence }& \ab{0.96 [0.84, 1.11] }& \ab{1.08 [1.05, 1.11] }& \ab{0.89 [0.77, 1.03] }& \ab{0.209}\\
\ab{Disclaimer }& \ab{1.47 [1.33, 1.64] }& \ab{1.67 [1.51, 1.84] }& \ab{0.89 [0.78, 1.00] }& \ab{0.209}\\
\ab{Refusal }& \ab{1.83 [1.59, 2.10] }& \ab{2.09 [1.82, 2.39] }& \ab{0.88 [0.73, 1.04] }& \ab{0.209}\\
\ab{Verification prompt }& \ab{1.23 [1.18, 1.28] }& \ab{1.24 [1.19, 1.29] }& \ab{1.00 [0.95, 1.05] }& \ab{0.891}\\
\ab{Reasoning trace }& \ab{1.21 [0.90, 1.63] }& \ab{1.28 [1.24, 1.33] }& \ab{0.94 [0.70, 1.27] }& \ab{0.787}\\
\bottomrule
\end{tabular}
\end{table}

\begin{table}[htbp]\centering\small
\caption{\ab{Observed qualifying-pair counts for selected feature-positive indicators used to interpret sparse direct contrasts. Other counts are not reported in the supplied summary exports.}}
\label{tab:pooled_counts}
\begin{tabular}{lr}\toprule \ab{Feature }& \ab{Detected qualifying pairs }\\\midrule
\ab{URL }& \ab{4,889}\\
\ab{Inspectable evidence }& \ab{5,372}\\
\ab{Broad attribution }& \ab{117,715}\\
\ab{Markdown link }& \ab{1,133}\\
\ab{Refusal }& \ab{2,159}\\
\ab{Uncertainty }& \ab{159,429}\\
\bottomrule\end{tabular}\end{table}

\begin{table}[H]\centering\small
\caption{Sensitivity to coarser platform fixed effects for sparse link-based indicators.}
\begin{tabular}{lrr}\toprule Feature & Checking RRR [95\% CI] & Challenging RRR [95\% CI]\\\midrule
URL & 1.73 [1.51, 1.99] & 1.86 [1.68, 2.07]\\
Inspectable evidence & 1.66 [1.47, 1.87] & 1.77 [1.60, 1.95]\\\bottomrule
\end{tabular}\end{table}

\section{Preserved source fields and feature overlap}\label{app:platform}
\ab{The platform-level prevalence heatmap is Figure~2B in the main manuscript; this section reports export-field availability and feature overlap.}

The source-release counts and platform-specific metadata fields are reported in Supplementary Section~A. Perplexity retains \texttt{source} and \texttt{source\_bar}, and Grok retains \texttt{links} and \texttt{source}; ChatGPT and Gemini do not preserve directly equivalent objects in the inspected export. These are schema descriptions; the difference cannot be interpreted as evidence that two products never displayed citations. In Perplexity, narrow citation-like attribution appears in 10.3\% of assistant turns, and 39.8\% of flagged turns lack a preserved source. A separate flagged-positive human audit of the derived variable gives precision around 0.95.
\begin{figure}[H]\centering\includegraphics[width=0.86\linewidth]{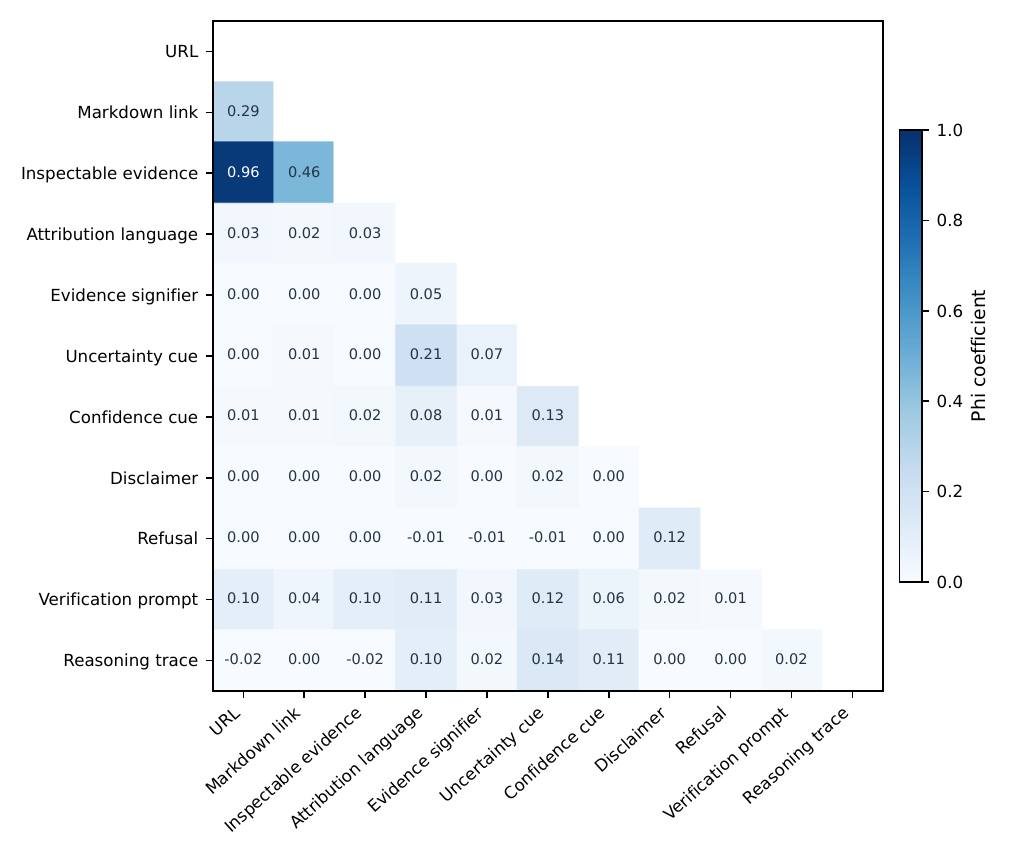}\caption{Pairwise $\phi$ between the 11 retained response features. URL and inspectable evidence are strongly nested.}\end{figure}

\paragraph{Interpreting preserved sources and signifiers.}
An explicit URL is a web address in the released answer; a markdown link encodes linked text; a preserved source object is stored in a platform-specific export field. The composite inspectable-evidence variable records the presence of at least one of these objects, not its accessibility or claim-level support. Broad attribution and evidence signifiers are text-pattern measures, so a response can refer to research without exposing a source. The narrower citation-like attribution flag is used for the source-preservation audit and should not be confused with broad attribution prevalence. Across the two datasets, these distinctions concern what the exported transcripts preserve, not what every user necessarily saw on screen.

\section{Extended sequential analyses}\label{app:sequence}
\ab{The supplementary sequence diagnostics use strictly adjacent turn transitions and within-conversation shuffles, whereas the fixed-effect lag analysis advances between consecutive qualifying exchanges and may skip intervening nonqualifying replies. The two designs cannot be substituted for each other. The reported shuffle figures are composition-preserving benchmarks; a full distribution of permutation differences was not retained in the provided summary, and no permutation significance test is claimed for these excesses. The split-panel jackknife is a sensitivity analysis: conversations close to the five-qualifying-pair threshold contribute only two or three observations per half-panel, limiting small-sample precision. Descriptive estimates for scrutiny-heavy conversations are restricted to conversations with at least five qualifying exchanges.}
\paragraph{\ab{Split-panel jackknife specification.}}
\ab{For a lag coefficient estimated in a full qualifying panel and in its two time-ordered halves, the reported bias-corrected estimate is}
\begin{equation}
\ab{\widehat\rho_{\mathrm{SPJ}}=2\widehat\rho_{\mathrm{FE,full}}-}
\ab{\frac{\widehat\rho_{\mathrm{FE,first}}+\widehat\rho_{\mathrm{FE,second}}}{2}.}
\end{equation}
\ab{The bootstrap resamples conversations, but cannot eliminate the instability of half-panels near the minimum five-pair threshold.}

\begin{table}[H]\centering\small
\caption{Strictly adjacent turn transitions, row percentages; all four outcomes retained.}
\begin{tabular}{lrrrr}\toprule
Previous turn & Accept & Check & Challenge & Not qualifying\\\midrule
Accept & 64.8 & 8.2 & 10.1 & 16.9\\
Check & 43.8 & 21.9 & 17.4 & 16.9\\
Challenge & 39.9 & 10.4 & 34.6 & 15.0\\
Not qualifying & 39.9 & 6.5 & 9.4 & 44.2\\\bottomrule
\end{tabular}\end{table}
\begin{table}[H]\centering\small
\caption{Same-type repetition: excess over transition from accepting, with within-conversation shuffle benchmarks (percentage points).}
\begin{tabular}{lrrr}\toprule
Transition & Observed excess & Shuffled excess & Extra ordered repetition\\\midrule
Check $\to$ check & 13.64 & 10.8 & $\approx2.8$\\
Challenge $\to$ challenge & 24.56 & 18.3 & $\approx6.3$\\\bottomrule
\end{tabular}\end{table}
Across 2,000 whole-conversation bootstrap resamples, the difference between the two observed excesses is 10.92 percentage points (95\% CI [10.1, 11.8]). In the 18,847 conversations with at least five qualifying exchanges, the top-decile scrutiny share threshold is 70.63\% (1,885 conversations). The scrutiny-heavy fraction is 14.0\% in Technical Help, 12.4\% in Seeking Information, and 5.2\% in Writing, without task adjustment. The mean longest consecutive scrutiny streak is 2.05 turns, compared with 1.88 under within-conversation permutation; 2.1\% of eligible conversations exceed their own 95th-percentile permutation benchmark. This suggests that most conversations do not show extraordinary extended scrutiny bursts relative to their own scrutiny counts.

\section{Complete response-feature codebook}\label{app:features}
Indicators are defined at the response level. Structural extraction indicates what the released artifact preserves. Linguistic measures detect surface expressions, not whether the model was calibrated or factually correct.

\paragraph{Development and scope of the detector families.}
\ab{The response-feature lexicons were constructed in two stages. The first drew candidate expressions from research on hedging, modality, epistemic stance, evidential language, refusals, and explanation. The second inspected matches in ShareChat and WildChat and refined expressions that gave predictable false positives. A bare instance of ``think,'' for example, was removed as an uncertainty marker because it often describes an action or introduces an opinion without expressing doubt. First-person expressions of uncertainty were distinguished during development from proposition-level uncertainty, although the pooled response-feature models use their broader parent indicator. Referrals to a professional or outside source were distinguished from disclaimers, which describe limits or conditions on the answer. All lexicons were frozen before the reported outcome models, and the human validation set was inspected after that freeze. The retained documentation does not provide the exact matching-corpus size, the number of pilot examples reviewed for each feature, or the number of people involved in iterative pattern refinement. We therefore do not assign numerical development-sample sizes; the separately documented final positive-case human audit is described in Supplementary Section~F.}

\paragraph{Broad attribution, citation-like attribution, and preserved sources.}
Broad attribution records language invoking research, experts, reports, or other external authority even when the assistant gives no identifiable source. Narrow citation-like attribution is a stricter pattern family intended for auditing whether an apparent source reference is accompanied by preserved support. The latter is not a twelfth primary regression feature. The source-preservation diagnostic flags narrow citation-like attribution with no URL, markdown link, or corresponding preserved platform source object. Perplexity's \texttt{source} and \texttt{source\_bar} fields, and Grok's \texttt{links} and \texttt{source} fields, are inspected where available; the common cross-corpus measures cannot assume identical export schemas. Citation-like language without a preserved source is a statement about the exported record and does not prove that a source was absent from the original user interface. The platform audit and feature-overlap matrix appear in Supplementary Section~C.

\paragraph{Binary response-level measurement and dependence.}
A detector is positive if a relevant expression or structure is found anywhere in the response or its preserved fields. This unit facilitates comparison across conversation corpora but cannot evaluate whether every claim is supported, whether a linked page is accessible, or whether its contents are reliable. The inspectable-evidence indicator is a union of potentially inspectable material, so explicit URLs and formatted links are nested within it. The strongest overlap among the retained measures is URL with inspectable evidence ($\phi=0.957$). A proposed broader evidentiary-language feature had near-perfect overlap with attribution ($\phi=0.995$) and was removed. To avoid treating highly overlapping cues as separable controlled exposures, the inferential models use a single focal response feature with a common adjustment set.

\paragraph{Illustrative coding examples.}
The following are conceptual illustrations of the coding categories, not quotations from analyzed conversations or a complete listing of the lexicon patterns. ``Research suggests'' illustrates broad attribution; ``the evidence shows'' illustrates an evidence signifier; ``may be'' illustrates uncertainty; ``I am certain'' illustrates confidence. A caution that an answer has limitations illustrates a disclaimer, a declined request illustrates refusal, and a suggestion to consult a primary source illustrates verification-prompt language. An assistant may exhibit multiple features in one response. A source URL or markdown link is a separate structural indicator, and exported thinking-field text is counted as a reasoning trace only when such a field is preserved. The codebook does not verify whether any cited source actually supports the response.

\begin{longtable}{p{0.18\textwidth}p{0.20\textwidth}p{0.30\textwidth}p{0.22\textwidth}}
\caption{Detailed operationalization of Study 1 response features.}\label{tab:study1_feature_codebook}\\
\toprule
\textbf{Role} & \textbf{Feature} & \textbf{Operational meaning} & \textbf{Interpretive caveat} \\
\midrule
\endfirsthead
\toprule
\textbf{Role} & \textbf{Feature} & \textbf{Operational meaning} & \textbf{Interpretive caveat} \\
\midrule
\endhead
Material verification support & URL & Explicit retrievable URL in the assistant response. & A link can be present without supporting the focal claim. \\
Material verification support & Markdown link & Link encoded in markdown structure. & A formatting signal; not a quality judgment. \\
Material verification support & Inspectable evidence & Composite indicator for response-level link or preserved source material that can potentially be opened or inspected. & Shared transcripts may omit platform-native panels. \\
Verification signifier & Attribution language & Language invoking studies, research, reports, experts, or other external authority. & Attribution can appear without retrievable support. \\
Verification signifier & Evidence signifier & Wording that presents a claim as evidenced or documented. & Signals evidentiary status without necessarily exposing evidence. \\
Epistemic communication cue & Uncertainty & Language marking uncertainty, qualification, or limited confidence. & Surface expression does not establish calibration. \\
Epistemic communication cue & Confidence & Strong commitment, certainty, or authoritative phrasing. & Surface confidence does not establish correctness. \\
Epistemic communication cue & Disclaimer & Statement about scope, limitations, caution, or conditions on the answer. & Can be informative or boilerplate. \\
Epistemic communication cue & Refusal & Assistant declines to provide the requested content or answer. & Refusals may coexist with alternatives or referrals. \\
Verification pathway & Prompt language & Language that suggests checking, consulting a source, or seeking external confirmation. & A suggested path may still impose substantial effort. \\
Epistemic communication cue & Reasoning trace & Preserved thinking, analysis, or process text in the shared transcript. & Process text does not provide external evidence. \\
\bottomrule
\end{longtable}
\section{Human precision audits and classification stability}\label{app:validation}

\paragraph{Distinct measurement-validity questions.}
\ab{The four-way outcome codebook defines an observable speech act at the next user turn. Its \emph{content/construct validity} depends on whether requesting confirmation and asserting an error correspond to the intended theoretical categories; the precedence rule resolves messages that combine both acts. Its \emph{human coding reliability} would require independent annotators applying that rule to the same cases, followed by an agreement statistic and adjudication. No such double-coded sample is documented, so human--human reliability remains unestimated. The \emph{classifier precision} below answers a narrower question: how often one blinded coder judged a label correct among messages to which the model assigned that label. Without model-negative sampling, recall and the complete checking-versus-challenging confusion matrix also remain unestimated. The audit documents useful classifier accuracy against one human interpretation, while independent replication of the coding boundary remains a priority for future validation.}

The final audits were conducted after freezing the production outcome codebook and response-feature lexicons, with model labels hidden from a single human coder. Sampling was conditional on the model's positive category assignment. The results are estimates of class-specific precision; they do not identify recall, F1, or population-level accuracy. The validation exercises are distinct from the earlier iterative pilot cases that informed codebook development.

\paragraph{Definition and denominator of scrutiny.}
We use \emph{scrutiny} throughout the manuscript for the union of checking and challenging. Among qualifying assistant--user response pairs, its prevalence is $(N_{\mathrm{checking}}+N_{\mathrm{challenging}})/(N_{\mathrm{accepting}}+N_{\mathrm{checking}}+N_{\mathrm{challenging}})$. ``Not qualifying'' is excluded from that denominator. This outcome records whether a user visibly requested confirmation or disputed an answer, not whether the user independently verified the information, was correct, or trusted the assistant. The four labels are mutually exclusive under the precedence rules below.

\paragraph{Boundary rules for the next-turn labels.}
The outcome codebook first asks whether the preceding assistant turn makes a checkable claim or recommendation and whether the next user message engages it. Without that condition, the pair is not qualifying. A request for the grounds or validity of a statement, such as ``What source supports this?'' or ``Are you sure?'', is checking when the user does not advance a competing answer. A reply that asserts an error, supplies a correction, or proposes a concrete alternative is challenging, regardless of whether the user's correction is ultimately right. Requests for a procedural explanation can be accepting if they build on the answer without questioning its validity. Mere acknowledgment is not necessarily accepting: context-free responses such as a bare ``ok'' or ``no'' can be too ambiguous to classify. Where a user both requests sources and asserts a correction, challenging takes precedence. These rules identify visible conversational moves; they do not infer tone, private trust, or successful verification beyond the chat.

\paragraph{Audit interpretation.}
\ab{Outcome-audit examples were sampled conditional on the label produced by the frozen classifier: 100 positives in each of the three qualifying classes from each corpus, with 50 additional not-qualifying cases. The positive-case design makes the reported fractions class-specific precision estimates. It does not sample model negatives in a way that supports recall or a confusion matrix for the population. The same distinction applies to the approximately 20 flagged responses inspected for each lexical detector. Because one human coder conducted the review, the audit does not estimate inter-annotator reliability. The absence of a second independent coder leaves uncertainty about reproducibility of the human judgments, in addition to the sampling uncertainty in the small positive-case detector audits. The 50-case audit of citation-like attribution without a source checks the derived flag against the preserved artifact, not the interface the user originally saw.}

\paragraph{Validation required to establish human reproducibility.}
\ab{The next validation step would draw a stratified sample of classifier-positive and classifier-negative response pairs from both corpora, deliberately including mixed requests, corrections phrased as questions, and short acknowledgments near the qualifying gate. At least two independent human annotators would apply the frozen four-class codebook without access to the model predictions. The evaluation would report an agreement statistic for the four outcomes, a focused checking-versus-challenging agreement analysis, and adjudicated boundary disagreements. Against the adjudicated labels, it would then report the four-class confusion matrix and class-specific precision and recall, including how often a true challenge is mislabeled as checking. This describes a proposed validation design; no such independent annotation or recall evaluation has been completed for the present study.}

\begin{table}[H]\centering
\caption{Outcome-label human audit, $N=650$ assistant--user pairs.}
\begin{tabular}{lrr}\toprule Outcome & Audited $N$ & Precision\\\midrule
Accepting & 200 & 0.81\\
Checking & 200 & 0.89\\
Challenging & 200 & 0.92\\
Not qualifying & 50 & 0.92\\\bottomrule
\end{tabular}\end{table}
For the qualifying outcome categories, 300 cases were drawn from ShareChat and 300 from WildChat, with 100 positive cases per category in each corpus. The other 50 cases assess the not-qualifying category. No statistically significant corpus difference in precision was detected. The production outcome labeler was Qwen3-30B-A3B, running with a frozen prompt and deterministic decoding configuration.

\begin{table}[H]\centering
\caption{Human precision of topic and intent controls in a 600-message first-turn audit.}
\begin{tabular}{llr}\toprule Level & Category & Precision\\\midrule
Intent & Asking & 0.80\\ & Doing & 0.85\\ & Expressing & 0.81\\\midrule
Coarse topic & Writing & 0.82\\ & Practical Guidance & 0.83\\ & Seeking Information & 0.91\\ & Technical Help & 0.95\\ & Multimedia & 0.90\\ & Self-Expression & 0.87\\ & Other/Unknown & 0.85\\\bottomrule
\end{tabular}\end{table}
The production topic-and-intent classifier was Qwen3-235B-A22B-Instruct-2507. A separate 200-conversation comparison of GPT-5.4-nano, Qwen3-235B, and DeepSeek-V4 found pairwise Cohen's $\kappa$ between 0.75 and 0.80 for intent, 0.58 and 0.69 for seven-category coarse topic, and 0.54 and 0.62 for the 24-category fine taxonomy. Cross-model agreement does not substitute for the human audit.

\begin{table}[H]\centering
\caption{Manual positive-case audits of frozen response-feature lexicons (approximately 20 flagged cases per detector).}
\begin{tabular}{lr}\toprule Detector & Precision\\\midrule
Broad attribution language & 0.67\\
Narrow citation-like attribution & 0.70\\
Evidence signifier & 0.68\\
Uncertainty cue & 0.72\\
Confidence cue & 0.85\\
Disclaimer & 0.82\\
Refusal & 0.75\\
Verification-prompt language & 0.81\\
First-person uncertainty & 0.70\\
Referral & 0.72\\\bottomrule
\end{tabular}\end{table}
\paragraph{Lexicon precision and prevalence sensitivity.}
Because each lexicon precision estimate comes from approximately 20 flagged examples, a few disagreements can materially affect the estimated value. These audits are useful for assessing measurement uncertainty, but the estimates should not be interpreted as precise rankings of detectors. For broad attribution, the estimated precision is 0.67. Applying it illustratively to the detected attribution prevalence of 31.9\% yields approximately 21.4\% detected true positives, or a 16:1 ratio relative to the observed 1.33\% explicit-URL rate. Allowing an illustrative precision range of roughly 0.43--0.85 gives an adjusted detected share of about 14--27\% and a ratio of about 10:1--20:1. These are sensitivity calculations, not confidence intervals for population prevalence. Positive-case precision audits do not identify false negatives or recall, and the comparison cannot correct for omitted information in platform exports.

A separate audit of 50 flagged cases of citation-like attribution without a preserved source yielded approximately 0.95 precision. It evaluates the existence of attribution in the absence of a preserved source object, URL, or markdown link. It does not establish that the original interface lacked a visible source. URL strings, markdown links, and preserved platform field objects were measured through structural rules, without a separate semantic precision audit.

\begin{landscape}
\section{Full task- and corpus-stratified estimates}\label{app:rq3}
\ab{This table preserves all attempted RQ3 strata, including non-estimable and unstable cells. For substantive claims in the main paper, we prioritize the conservative 48-test non-Multimedia family, treating 14 non-estimable contrasts as $p=1$ and carrying forward two finite-but-unstable values. The 32 stable-contrast correction is retained as a secondary sensitivity family.}
\begin{table}[htbp]
\centering\scriptsize
\caption{\ab{RQ3 multiplicity families. Component coefficients compare each outcome with acceptance; direct contrasts compare checking with challenging.}}
\label{tab:rq3-families}
\setlength{\tabcolsep}{3.5pt}
\begin{tabular}{p{.24\linewidth}p{.39\linewidth}rp{.18\linewidth}}
\toprule
\ab{Family }& \ab{Scope }& \ab{Tests }& \ab{Reported in }\\\midrule
\ab{Component coefficients }& \ab{Nine strata $\times$ six features $\times$ two outcomes; BH-FDR }& \ab{108 }& \ab{Main Figure~4A; Section~G }\\
\ab{All planned direct }& \ab{Nine strata $\times$ six features, including Multimedia }& \ab{54 }& \ab{Supp.~G }\\
\ab{Primary direct }& \ab{Stable non-Multimedia contrasts; BH-FDR }& \ab{32 }& \ab{Main Figure~4B; Section~G }\\
\ab{Conservative direct }& \ab{Non-Multimedia: 32 stable, two unstable, 14 NE ($p=1$); BH-FDR }& \ab{48 }& \ab{Main Figure~4B; Section~G }\\\bottomrule
\end{tabular}
\end{table}

\scriptsize
\setlength{\tabcolsep}{3pt}
\begin{longtable}{lllrrrl}
\caption{\ab{All S1-RQ3 stratified checking-versus-challenging contrasts. NE denotes a non-estimable or unstable specification. The $q$ column reports the estimable-only correction; conservative $q$ values for the significant contrasts appear in the text and Figure~4 of the main manuscript.}}\label{tab:study1_rq3_full}\\
\toprule
\textbf{Type} & \textbf{Stratum} & \textbf{Feature} & \textbf{Check RRR [95\% CI]} & \textbf{Challenge RRR [95\% CI]} & \textbf{Check/Challenge [95\% CI]} & \textbf{FDR $q$ (32 stable)} \\
\midrule
\endfirsthead
\toprule
\textbf{Type} & \textbf{Stratum} & \textbf{Feature} & \textbf{Check RRR [95\% CI]} & \textbf{Challenge RRR [95\% CI]} & \textbf{Check/Challenge [95\% CI]} & \textbf{FDR $q$ (32 stable)} \\
\midrule
\endhead
Topic & Multimedia & URL & NE & NE & NE & -- \\
Topic & Multimedia & Inspectable evidence & NE & NE & NE & -- \\
Topic & Multimedia & Uncertainty & NE & NE & NE & -- \\
Topic & Multimedia & Reasoning trace & NE & NE & NE & -- \\
Topic & Multimedia & Disclaimer & NE & NE & NE & -- \\
Topic & Multimedia & Refusal & NE & NE & NE & -- \\
Topic & Other/Unknown & URL & 1.56 [1.12, 2.17] & 1.58 [1.12, 2.23] & 0.99 [0.70, 1.39] & 0.963 \\
Topic & Other/Unknown & Inspectable evidence & 1.44 [1.04, 1.98] & 1.47 [1.06, 2.03] & 0.98 [0.70, 1.38] & 0.963 \\
Topic & Other/Unknown & Uncertainty & 1.10 [1.03, 1.19] & 1.12 [1.04, 1.19] & 0.99 [0.91, 1.08] & 0.963 \\
Topic & Other/Unknown & Reasoning trace & 1.24 [1.14, 1.36] & 1.37 [1.26, 1.49] & 0.91 [0.82, 1.00] & 0.178 \\
Topic & Other/Unknown & Disclaimer & 1.61 [1.29, 2.00] & 1.97 [1.64, 2.38] & 0.82 [0.64, 1.04] & 0.234 \\
Topic & Other/Unknown & Refusal & 2.14 [1.61, 2.85] & 2.20 [1.72, 2.83] & 0.97 [0.72, 1.31] & 0.963 \\
Topic & Practical Guidance & URL & 1.70 [1.15, 2.50] & 2.16 [1.33, 3.49] & 0.79 [0.38, 1.62] & 0.746 \\
Topic & Practical Guidance & Inspectable evidence & NE & NE & NE & -- \\
Topic & Practical Guidance & Uncertainty & 0.99 [0.91, 1.09] & 1.04 [0.93, 1.16] & 0.96 [0.84, 1.08] & 0.746 \\
Topic & Practical Guidance & Reasoning trace & NE & NE & NE & -- \\
Topic & Practical Guidance & Disclaimer & 1.34 [0.90, 1.99] & 1.16 [0.76, 1.76] & 1.16 [0.70, 1.91] & 0.785 \\
Topic & Practical Guidance & Refusal & NE & NE & NE & -- \\
Topic & Seeking Information & URL & 1.25 [0.99, 1.58] & 1.23 [0.96, 1.58] & 1.01 [0.77, 1.33] & 0.963 \\
Topic & Seeking Information & Inspectable evidence & 1.17 [0.92, 1.48] & 1.19 [0.92, 1.53] & 0.98 [0.76, 1.27] & 0.963 \\
Topic & Seeking Information & Uncertainty & 1.08 [1.03, 1.13] & 1.08 [1.03, 1.14] & 1.00 [0.94, 1.06] & 0.963 \\
Topic & Seeking Information & Reasoning trace & 1.23 [1.16, 1.30] & 1.38 [1.30, 1.47] & 0.89 [0.83, 0.95] & 0.007 \\
Topic & Seeking Information & Disclaimer & 1.37 [1.16, 1.62] & 1.66 [1.41, 1.95] & 0.83 [0.67, 1.02] & 0.216 \\
Topic & Seeking Information & Refusal & 1.83 [1.46, 2.29] & 2.93 [2.32, 3.71] & 0.62 [0.47, 0.83] & 0.010 \\
Topic & Self-Expression & URL & NE & NE & NE & -- \\
Topic & Self-Expression & Inspectable evidence & 1.46 [0.56, 3.83] & 0.91 [0.43, 1.90] & 1.61 [0.74, 3.50] & 0.456 \\
Topic & Self-Expression & Uncertainty & 1.07 [0.96, 1.20] & 1.02 [0.92, 1.14] & 1.05 [0.92, 1.19] & 0.746 \\
Topic & Self-Expression & Reasoning trace & NE & NE & NE & -- \\
Topic & Self-Expression & Disclaimer & NE & NE & NE & -- \\
Topic & Self-Expression & Refusal & NE & NE & NE & -- \\
Topic & Technical Help & URL & 1.62 [1.39, 1.90] & 1.43 [1.26, 1.63] & 1.14 [0.96, 1.34] & 0.300 \\
Topic & Technical Help & Inspectable evidence & NE & NE & NE & -- \\
Topic & Technical Help & Uncertainty & 1.08 [1.01, 1.14] & 0.86 [0.81, 0.91] & 1.26 [1.17, 1.34] & $<0.001$ \\
Topic & Technical Help & Reasoning trace & NE & NE & NE & -- \\
Topic & Technical Help & Disclaimer & 0.93 [0.65, 1.34] & 0.92 [0.67, 1.26] & 1.02 [0.68, 1.53] & 0.963 \\
Topic & Technical Help & Refusal & 1.30 [0.83, 2.05] & 0.90 [0.64, 1.26] & 1.45 [0.88, 2.40] & 0.317 \\
Topic & Writing & URL & NE & NE & NE & -- \\
Topic & Writing & Inspectable evidence & NE & NE & NE & -- \\
Topic & Writing & Uncertainty & 1.18 [1.10, 1.26] & 1.09 [1.03, 1.17] & 1.08 [0.99, 1.17] & 0.216 \\
Topic & Writing & Reasoning trace & 1.35 [1.24, 1.48] & 1.50 [1.39, 1.62] & 0.90 [0.81, 1.00] & 0.178 \\
Topic & Writing & Disclaimer & NE & NE & NE & -- \\
Topic & Writing & Refusal & NE & NE & NE & -- \\
Corpus & ShareChat & URL & 1.15 [0.87, 1.50] & 1.29 [1.01, 1.66] & 0.89 [0.63, 1.25] & 0.746 \\
Corpus & ShareChat & Inspectable evidence & 1.10 [0.92, 1.32] & 1.20 [0.96, 1.49] & 0.92 [0.71, 1.18] & 0.746 \\
Corpus & ShareChat & Uncertainty & 1.07 [1.02, 1.12] & 1.07 [1.04, 1.11] & 1.00 [0.94, 1.06] & 0.963 \\
Corpus & ShareChat & Reasoning trace & 1.32 [1.23, 1.41] & 1.50 [1.45, 1.56] & 0.88 [0.82, 0.94] & 0.004 \\
Corpus & ShareChat & Disclaimer & 1.40 [1.19, 1.63] & 1.77 [1.54, 2.03] & 0.79 [0.66, 0.94] & 0.048 \\
Corpus & ShareChat & Refusal & 1.87 [1.57, 2.23] & 2.36 [1.99, 2.79] & 0.79 [0.64, 0.98] & 0.162 \\
Corpus & WildChat & URL & NE & NE & NE & -- \\
Corpus & WildChat & Inspectable evidence & NE & NE & NE & -- \\
Corpus & WildChat & Uncertainty & 1.18 [0.93, 1.50] & 0.96 [0.92, 1.00] & 1.23 [0.97, 1.56] & 0.234 \\
Corpus & WildChat & Reasoning trace & 0.99 [0.91, 1.08] & 0.86 [0.81, 0.91] & 1.16 [1.05, 1.28] & 0.014 \\
Corpus & WildChat & Disclaimer & NE & NE & NE & -- \\
Corpus & WildChat & Refusal & 1.76 [1.34, 2.31] & 1.55 [1.26, 1.90] & 1.14 [0.84, 1.54] & 0.746 \\
\bottomrule
\end{longtable}
\end{landscape}

\begingroup
\scriptsize
\setlength{\tabcolsep}{3pt}
\setlength\LTleft{0pt}\setlength\LTright{0pt}
\refstepcounter{table}\label{tab:rq3_beta_full}\noindent\textbf{Table~\thetable. RQ3 coefficient tests: all 108 planned comparisons.}\par\smallskip
\begin{longtable}{p{2.2cm}p{2.5cm}p{2.2cm}r p{1.8cm}r r l}

\toprule
Stratum & Feature & Contrast & RRR & 95\% CI & AME (pp) & $q_{108}$ & Status \\
\midrule
\endfirsthead
\multicolumn{8}{l}{\textit{Table continued from previous page}}\\
\toprule
Stratum & Feature & Contrast & RRR & 95\% CI & AME (pp) & $q_{108}$ & Status \\
\midrule
\endhead
\bottomrule
\endfoot
Writing & URL & Check / Accept & -- & -- & -- & 1.000 & singular \\
\rowcolor{black!3}
Writing & URL & Challenge / Accept & -- & -- & -- & 1.000 & singular \\
Writing & Inspectable evidence & Check / Accept & -- & -- & -- & 1.000 & singular \\
\rowcolor{black!3}
Writing & Inspectable evidence & Challenge / Accept & -- & -- & -- & 1.000 & singular \\
Writing & Uncertainty & Check / Accept & 1.18 & [1.10, 1.26] & 1.11 & <0.001 & estimable \\
\rowcolor{black!3}
Writing & Uncertainty & Challenge / Accept & 1.09 & [1.03, 1.17] & 0.81 & 0.017 & estimable \\
Writing & Reasoning trace & Check / Accept & 1.35 & [1.24, 1.48] & 1.79 & <0.001 & estimable \\
\rowcolor{black!3}
Writing & Reasoning trace & Challenge / Accept & 1.50 & [1.39, 1.62] & 4.49 & <0.001 & estimable \\
Writing & Disclaimer & Check / Accept & -- & -- & -- & 1.000 & singular \\
\rowcolor{black!3}
Writing & Disclaimer & Challenge / Accept & -- & -- & -- & 1.000 & singular \\
Writing & Refusal & Check / Accept & -- & -- & -- & 1.000 & singular \\
\rowcolor{black!3}
Writing & Refusal & Challenge / Accept & -- & -- & -- & 1.000 & singular \\
Practical Guidance & URL & Check / Accept & 1.70 & [1.15, 2.50] & 4.64 & 0.023 & estimable \\
\rowcolor{black!3}
Practical Guidance & URL & Challenge / Accept & 2.16 & [1.33, 3.49] & 8.23 & 0.006 & estimable \\
Practical Guidance & Inspectable evidence & Check / Accept & -- & -- & -- & 1.000 & nonfinite \\
\rowcolor{black!3}
Practical Guidance & Inspectable evidence & Challenge / Accept & -- & -- & -- & 1.000 & nonfinite \\
Practical Guidance & Uncertainty & Check / Accept & 0.99 & [0.91, 1.09] & -0.13 & 1.000 & estimable \\
\rowcolor{black!3}
Practical Guidance & Uncertainty & Challenge / Accept & 1.04 & [0.93, 1.16] & 0.39 & 0.885 & estimable \\
Practical Guidance & Reasoning trace & Check / Accept & -- & -- & -- & 1.000 & singular \\
\rowcolor{black!3}
Practical Guidance & Reasoning trace & Challenge / Accept & -- & -- & -- & 1.000 & singular \\
Practical Guidance & Disclaimer & Check / Accept & 1.34 & [0.90, 1.99] & 3.12 & 0.331 & estimable \\
\rowcolor{black!3}
Practical Guidance & Disclaimer & Challenge / Accept & 1.16 & [0.76, 1.76] & 0.96 & 0.922 & estimable \\
Practical Guidance & Refusal & Check / Accept & -- & -- & -- & 1.000 & nonfinite \\
\rowcolor{black!3}
Practical Guidance & Refusal & Challenge / Accept & -- & -- & -- & 1.000 & nonfinite \\
Seeking Information & URL & Check / Accept & 1.25 & [0.99, 1.58] & 2.24 & 0.153 & estimable \\
\rowcolor{black!3}
Seeking Information & URL & Challenge / Accept & 1.23 & [0.96, 1.58] & 2.49 & 0.235 & estimable \\
Seeking Information & Inspectable evidence & Check / Accept & 1.17 & [0.92, 1.48] & 1.44 & 0.425 & estimable \\
\rowcolor{black!3}
Seeking Information & Inspectable evidence & Challenge / Accept & 1.19 & [0.92, 1.53] & 2.17 & 0.375 & estimable \\
Seeking Information & Uncertainty & Check / Accept & 1.08 & [1.03, 1.13] & 0.74 & 0.004 & estimable \\
\rowcolor{black!3}
Seeking Information & Uncertainty & Challenge / Accept & 1.08 & [1.03, 1.14] & 0.96 & 0.004 & estimable \\
Seeking Information & Reasoning trace & Check / Accept & 1.23 & [1.16, 1.30] & 1.61 & <0.001 & estimable \\
\rowcolor{black!3}
Seeking Information & Reasoning trace & Challenge / Accept & 1.38 & [1.30, 1.47] & 4.39 & <0.001 & estimable \\
Seeking Information & Disclaimer & Check / Accept & 1.37 & [1.16, 1.62] & 2.38 & <0.001 & estimable \\
\rowcolor{black!3}
Seeking Information & Disclaimer & Challenge / Accept & 1.66 & [1.41, 1.95] & 7.28 & <0.001 & estimable \\
Seeking Information & Refusal & Check / Accept & 1.83 & [1.46, 2.29] & 3.47 & <0.001 & estimable \\
\rowcolor{black!3}
Seeking Information & Refusal & Challenge / Accept & 2.93 & [2.32, 3.71] & 17.41 & <0.001 & estimable \\
Technical Help & URL & Check / Accept & 1.62 & [1.39, 1.90] & 4.67 & <0.001 & estimable \\
\rowcolor{black!3}
Technical Help & URL & Challenge / Accept & 1.43 & [1.26, 1.63] & 4.65 & <0.001 & estimable \\
Technical Help & Inspectable evidence & Check / Accept & -- & -- & -- & 1.000 & nonfinite \\
\rowcolor{black!3}
Technical Help & Inspectable evidence & Challenge / Accept & -- & -- & -- & 1.000 & nonfinite \\
Technical Help & Uncertainty & Check / Accept & 1.08 & [1.01, 1.14] & 1.30 & 0.052 & estimable \\
\rowcolor{black!3}
Technical Help & Uncertainty & Challenge / Accept & 0.86 & [0.81, 0.91] & -2.85 & <0.001 & estimable \\
Technical Help & Reasoning trace & Check / Accept & -- & -- & -- & 1.000 & nonfinite \\
\rowcolor{black!3}
Technical Help & Reasoning trace & Challenge / Accept & -- & -- & -- & 1.000 & nonfinite \\
Technical Help & Disclaimer & Check / Accept & 0.93 & [0.65, 1.34] & -0.48 & 1.000 & estimable \\
\rowcolor{black!3}
Technical Help & Disclaimer & Challenge / Accept & 0.92 & [0.67, 1.26] & -1.28 & 1.000 & estimable \\
Technical Help & Refusal & Check / Accept & 1.30 & [0.83, 2.05] & 3.59 & 0.513 & estimable \\
\rowcolor{black!3}
Technical Help & Refusal & Challenge / Accept & 0.90 & [0.64, 1.26] & -2.69 & 0.956 & estimable \\
Multimedia & URL & Check / Accept & -- & -- & -- & 1.000 & nonfinite \\
\rowcolor{black!3}
Multimedia & URL & Challenge / Accept & -- & -- & -- & 1.000 & nonfinite \\
Multimedia & Inspectable evidence & Check / Accept & -- & -- & -- & 1.000 & nonfinite \\
\rowcolor{black!3}
Multimedia & Inspectable evidence & Challenge / Accept & -- & -- & -- & 1.000 & nonfinite \\
Multimedia & Uncertainty & Check / Accept & -- & -- & -- & 1.000 & nonfinite \\
\rowcolor{black!3}
Multimedia & Uncertainty & Challenge / Accept & -- & -- & -- & 1.000 & nonfinite \\
Multimedia & Reasoning trace & Check / Accept & -- & -- & -- & 1.000 & nonfinite \\
\rowcolor{black!3}
Multimedia & Reasoning trace & Challenge / Accept & -- & -- & -- & 1.000 & nonfinite \\
Multimedia & Disclaimer & Check / Accept & -- & -- & -- & 1.000 & nonfinite \\
\rowcolor{black!3}
Multimedia & Disclaimer & Challenge / Accept & -- & -- & -- & 1.000 & nonfinite \\
Multimedia & Refusal & Check / Accept & -- & -- & -- & 1.000 & nonfinite \\
\rowcolor{black!3}
Multimedia & Refusal & Challenge / Accept & -- & -- & -- & 1.000 & nonfinite \\
Self-Expression & URL & Check / Accept & -- & -- & -- & 1.000 & nonfinite \\
\rowcolor{black!3}
Self-Expression & URL & Challenge / Accept & -- & -- & -- & 1.000 & nonfinite \\
Self-Expression & Inspectable evidence & Check / Accept & 1.46 & [0.56, 3.83] & 3.19 & 0.836 & estimable \\
\rowcolor{black!3}
Self-Expression & Inspectable evidence & Challenge / Accept & 0.91 & [0.43, 1.90] & -1.72 & 1.000 & estimable \\
Self-Expression & Uncertainty & Check / Accept & 1.07 & [0.96, 1.20] & 0.44 & 0.489 & estimable \\
\rowcolor{black!3}
Self-Expression & Uncertainty & Challenge / Accept & 1.02 & [0.92, 1.14] & 0.19 & 1.000 & estimable \\
Self-Expression & Reasoning trace & Check / Accept & -- & -- & -- & 1.000 & singular \\
\rowcolor{black!3}
Self-Expression & Reasoning trace & Challenge / Accept & -- & -- & -- & 1.000 & singular \\
Self-Expression & Disclaimer & Check / Accept & -- & -- & -- & 1.000 & nonfinite \\
\rowcolor{black!3}
Self-Expression & Disclaimer & Challenge / Accept & -- & -- & -- & 1.000 & nonfinite \\
Self-Expression & Refusal & Check / Accept & -- & -- & -- & 1.000 & nonfinite \\
\rowcolor{black!3}
Self-Expression & Refusal & Challenge / Accept & -- & -- & -- & 1.000 & nonfinite \\
Other/Unknown & URL & Check / Accept & 1.56 & [1.12, 2.17] & 3.71 & 0.024 & estimable \\
\rowcolor{black!3}
Other/Unknown & URL & Challenge / Accept & 1.58 & [1.12, 2.23] & 6.39 & 0.024 & estimable \\
Other/Unknown & Inspectable evidence & Check / Accept & 1.44 & [1.04, 1.98] & 2.99 & 0.066 & estimable \\
\rowcolor{black!3}
Other/Unknown & Inspectable evidence & Challenge / Accept & 1.47 & [1.06, 2.03] & 5.29 & 0.056 & estimable \\
Other/Unknown & Uncertainty & Check / Accept & 1.10 & [1.03, 1.19] & 0.77 & 0.024 & estimable \\
\rowcolor{black!3}
Other/Unknown & Uncertainty & Challenge / Accept & 1.12 & [1.04, 1.19] & 1.45 & 0.004 & estimable \\
Other/Unknown & Reasoning trace & Check / Accept & 1.24 & [1.14, 1.36] & 1.51 & <0.001 & estimable \\
\rowcolor{black!3}
Other/Unknown & Reasoning trace & Challenge / Accept & 1.37 & [1.26, 1.49] & 4.49 & <0.001 & estimable \\
Other/Unknown & Disclaimer & Check / Accept & 1.61 & [1.29, 2.00] & 3.21 & <0.001 & estimable \\
\rowcolor{black!3}
Other/Unknown & Disclaimer & Challenge / Accept & 1.97 & [1.64, 2.38] & 10.54 & <0.001 & estimable \\
Other/Unknown & Refusal & Check / Accept & 2.14 & [1.61, 2.85] & 6.48 & <0.001 & estimable \\
\rowcolor{black!3}
Other/Unknown & Refusal & Challenge / Accept & 2.20 & [1.72, 2.83] & 11.44 & <0.001 & estimable \\
ShareChat & URL & Check / Accept & 1.15 & [0.87, 1.50] & 0.81 & 0.631 & estimable \\
\rowcolor{black!3}
ShareChat & URL & Challenge / Accept & 1.29 & [1.01, 1.66] & 3.64 & 0.106 & estimable \\
ShareChat & Inspectable evidence & Check / Accept & 1.10 & [0.92, 1.32] & 0.60 & 0.564 & estimable \\
\rowcolor{black!3}
ShareChat & Inspectable evidence & Challenge / Accept & 1.20 & [0.96, 1.49] & 2.54 & 0.234 & estimable \\
ShareChat & Uncertainty & Check / Accept & 1.07 & [1.02, 1.12] & 0.55 & 0.027 & estimable \\
\rowcolor{black!3}
ShareChat & Uncertainty & Challenge / Accept & 1.07 & [1.04, 1.11] & 0.83 & <0.001 & estimable \\
ShareChat & Reasoning trace & Check / Accept & 1.32 & [1.23, 1.41] & 1.97 & <0.001 & estimable \\
\rowcolor{black!3}
ShareChat & Reasoning trace & Challenge / Accept & 1.50 & [1.45, 1.56] & 5.62 & <0.001 & estimable \\
ShareChat & Disclaimer & Check / Accept & 1.40 & [1.19, 1.63] & 2.06 & <0.001 & estimable \\
\rowcolor{black!3}
ShareChat & Disclaimer & Challenge / Accept & 1.77 & [1.54, 2.03] & 8.55 & <0.001 & estimable \\
ShareChat & Refusal & Check / Accept & 1.87 & [1.57, 2.23] & 4.53 & <0.001 & estimable \\
\rowcolor{black!3}
ShareChat & Refusal & Challenge / Accept & 2.36 & [1.99, 2.79] & 12.99 & <0.001 & estimable \\
WildChat & URL & Check / Accept & -- & -- & -- & 1.000 & unstable \\
\rowcolor{black!3}
WildChat & URL & Challenge / Accept & 1.39 & [1.24, 1.56] & 3.11 & <0.001 & estimable \\
WildChat & Inspectable evidence & Check / Accept & -- & -- & -- & 1.000 & unstable \\
\rowcolor{black!3}
WildChat & Inspectable evidence & Challenge / Accept & 1.39 & [1.24, 1.55] & 3.11 & <0.001 & estimable \\
WildChat & Uncertainty & Check / Accept & 1.18 & [0.93, 1.50] & 1.53 & 0.368 & estimable \\
\rowcolor{black!3}
WildChat & Uncertainty & Challenge / Accept & 0.96 & [0.92, 1.00] & -0.81 & 0.145 & estimable \\
WildChat & Reasoning trace & Check / Accept & 0.99 & [0.91, 1.08] & 0.20 & 1.000 & estimable \\
\rowcolor{black!3}
WildChat & Reasoning trace & Challenge / Accept & 0.86 & [0.81, 0.91] & -1.85 & <0.001 & estimable \\
WildChat & Disclaimer & Check / Accept & -- & -- & -- & 1.000 & nonfinite \\
\rowcolor{black!3}
WildChat & Disclaimer & Challenge / Accept & -- & -- & -- & 1.000 & nonfinite \\
WildChat & Refusal & Check / Accept & 1.76 & [1.34, 2.31] & 4.77 & <0.001 & estimable \\
\rowcolor{black!3}
WildChat & Refusal & Challenge / Accept & 1.55 & [1.26, 1.90] & 4.64 & <0.001 & estimable \\
\end{longtable}
\endgroup

\section{Held-out model comparison and exploratory model heterogeneity}\label{app:model}
\subsection{Predictive comparison}
The analysis below provides the full six-block comparison discussed in RQ3. Five-fold held-out evaluation uses folds grouped at the conversation level, so no pair from a held-out conversation appears in the training set. AUC is macro one-versus-rest across the three qualifying outcome labels, and log loss is multiclass. This exercise characterizes the predictive information in the observed variables, not the consequences of interventions on interface cues.
\begin{table}[H]\centering\small
\caption{In-sample and held-out model fit. Parentheses give the standard deviation over five grouped folds.}
\begin{tabular}{p{0.38\linewidth}rrr}
\toprule Predictor blocks & In-sample AUC & Held-out AUC & Held-out log loss\\\midrule
Response features only & 0.556 & 0.551 (0.004) & 0.788 (0.007)\\
Topic, intent, turn position & 0.594 & 0.592 (0.005) & 0.779 (0.008)\\
Task block + response features & 0.607 & 0.603 (0.004) & 0.775 (0.007)\\
Task block + model/system & 0.610 & 0.605 (0.004) & 0.774 (0.007)\\
Task + model/system + features & 0.620 & 0.615 (0.003) & 0.771 (0.007)\\
Full: preceding + length and truncation & 0.625 & 0.621 (0.003) & 0.769 (0.007)\\\bottomrule
\end{tabular}
\end{table}

\subsection{Exploratory reasoning-model and model-version comparisons}
These model-level comparisons are exploratory and are not part of the primary 48-test family. Within WildChat, scrutiny among qualifying responses varies from 20.1\% for \texttt{gpt-3.5-turbo-0125} to 34.8\% for \texttt{gpt-4o-2024-11-20}. Across 12 retained descriptive model groups, the correlation between scrutiny and the separately labeled reliance indicator is approximately $r=-0.90$; model-specific regression estimates were available for 11 of those groups. Much of this inverse association is mechanical because turn-level reliance and scrutiny are near-complements in the coding: the ecological correlation should not be interpreted as an independent model-level behavioral effect. In model-specific exploratory specifications, attribution is positively associated with scrutiny in four of 11 estimable models and never negatively; confidence is associated with less scrutiny in six of 11 and never with more. Uncertainty varies in direction; inspectable-evidence indicators do not reach significance in the model-specific specifications. These results also depend on model-group composition.

An additional ShareChat analysis compares ChatGPT responses with \emph{explicitly identified} reasoning and non-reasoning model IDs. Reasoning-model responses contain more uncertainty (46.7\% versus 37.1\%) but somewhat less inspectable support (1.36\% versus 1.87\%). Visible scrutiny follows 37.6\% of qualifying reasoning-model responses and 32.5\% of qualifying non-reasoning responses. The covariate-adjusted odds ratio is 1.31 (95\% CI [1.18, 1.45]). Because reasoning-model assignment was not randomized and model/version grouping is correlated with deployment era, task mix, and platform behavior, this result is descriptive. WildChat did not provide a comparable reasoning/non-reasoning model-ID split in the examined sample.

\section{Supplementary result figures and diagnostic plots}\label{app:all_figures}

The following supplementary figures document the distribution of response features, validation diagnostics, contextual heterogeneity, and sequential analyses. Figure captions specify the sampling frame and comparison.

The qualifying fraction is higher for ShareChat (80.4\%) than for WildChat (66.5\%), and the average number of response pairs per conversation differs across platforms (Figure~\ref{fig:new_sample}).

\begin{figure}[H]
\centering
\includegraphics[width=0.86\textwidth]{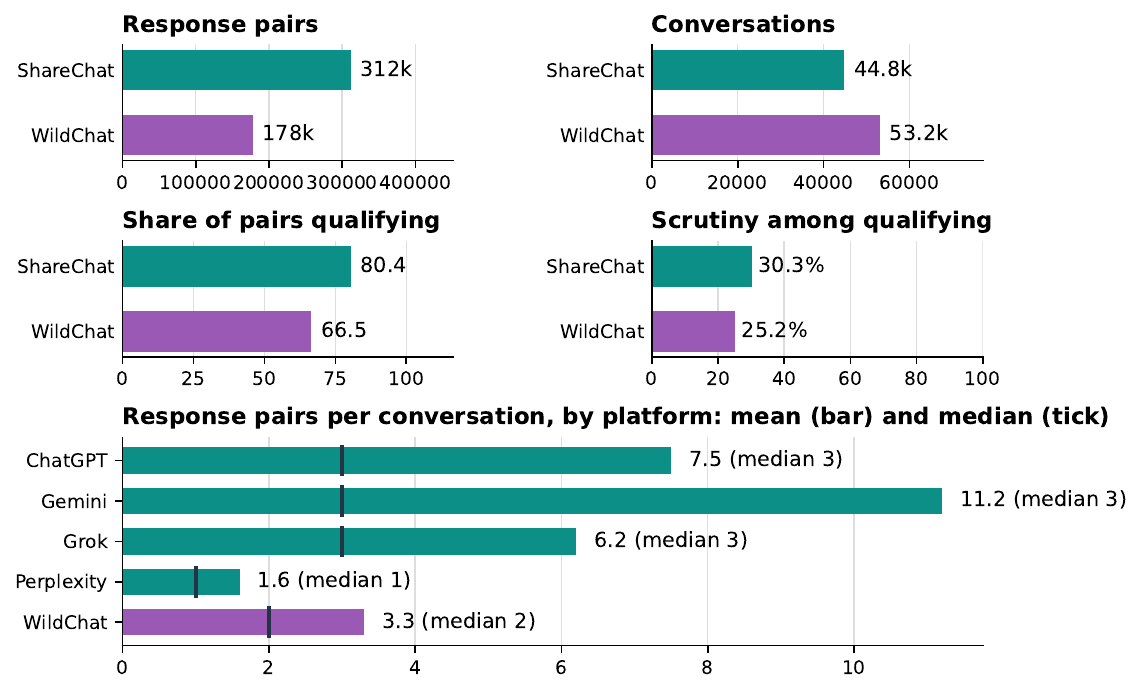}
\caption{Response-pair counts, conversation counts, qualifying rates, and platform-specific pair counts. Joined sample and platform denominators.}
\label{fig:new_sample}
\end{figure}

Audit precision varies across label families; outcome labels have 0.81--0.92 estimated precision, compared with 0.67--0.85 for the lexicons (Figure~\ref{fig:fig_extra_validation}).

\begin{figure}[H]
\centering
\includegraphics[width=0.86\textwidth]{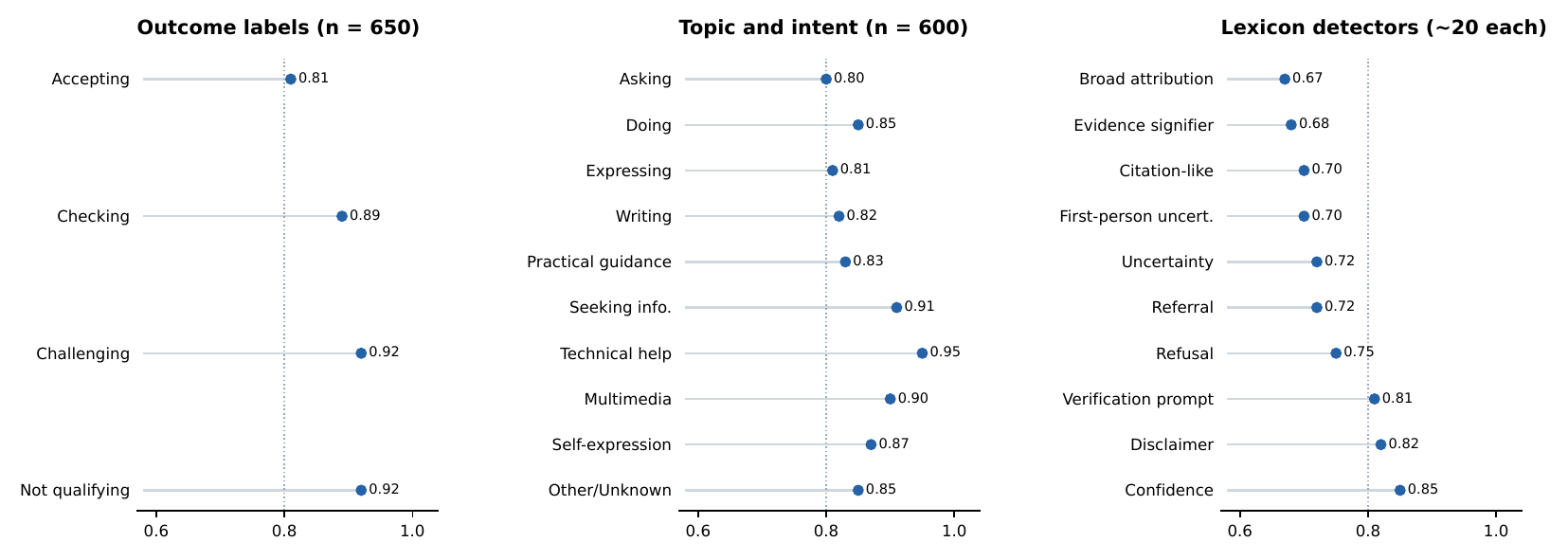}
\caption{Blinded audits of outcome, topic/intent, and lexicon precision. Positive-case audits estimate precision, not recall.}
\label{fig:fig_extra_validation}
\end{figure}

Attribution, uncertainty, and other response-level cues have different frequencies across systems (Figure~\ref{fig:new_platform_cues}).

\begin{figure}[H]
\centering
\includegraphics[width=0.86\textwidth]{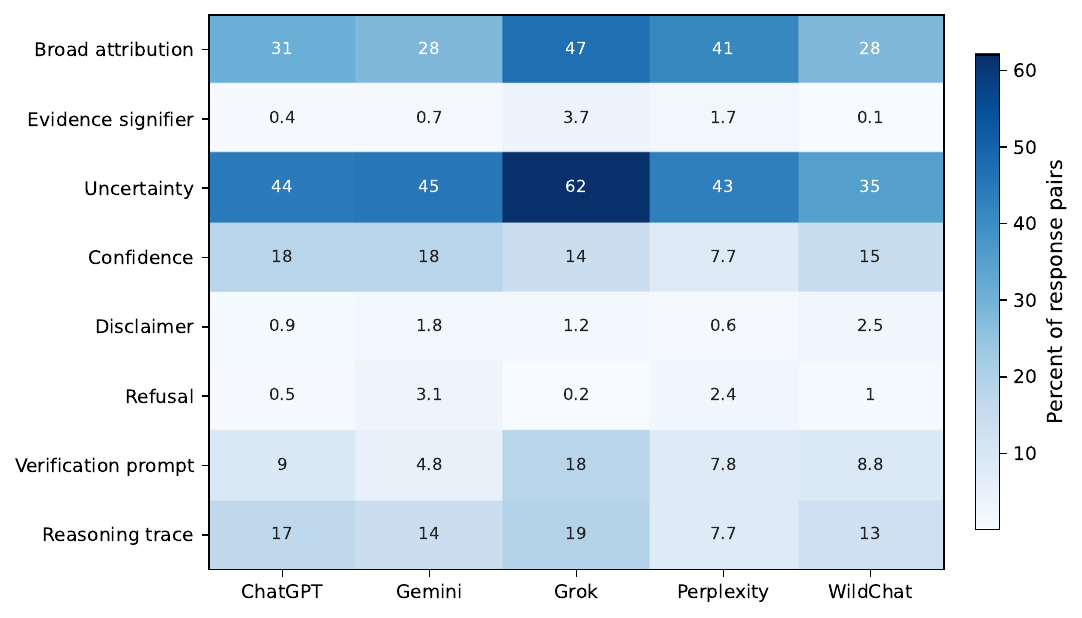}
\caption{Platform-specific frequency of epistemic and attribution cues in the full joined sample.}
\label{fig:new_platform_cues}
\end{figure}

URLs and inspectable evidence remain sparse in most platforms in the exported records (Figure~\ref{fig:new_platform_support}).

\begin{figure}[H]
\centering
\includegraphics[width=0.86\textwidth]{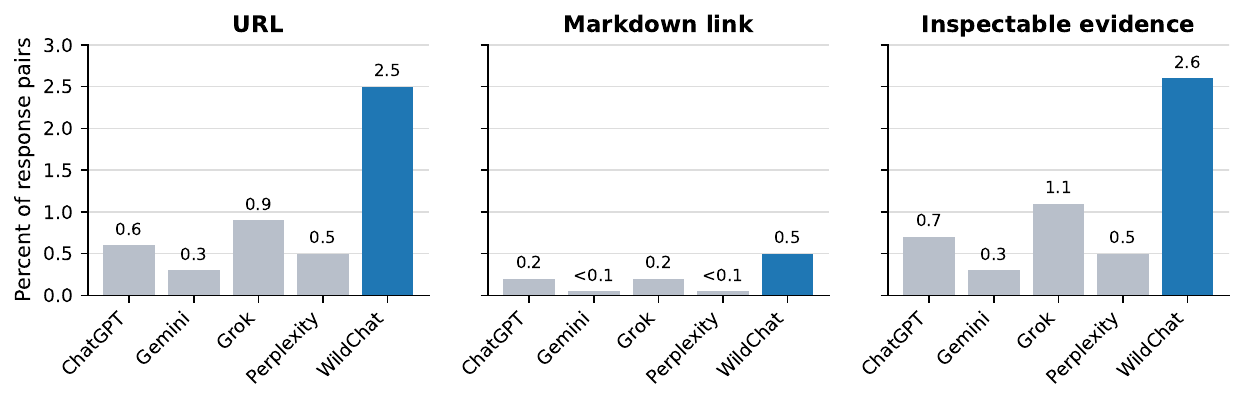}
\caption{Platform-specific URL and inspectable-support prevalence. Denominators are full joined response pairs.}
\label{fig:new_platform_support}
\end{figure}

Within Technical Help, uncertainty displays a notably different checking and challenging association (Figure~\ref{fig:new_rq3_tech}). Non-estimable coefficients are marked explicitly.

\begin{figure}[H]
\centering
\includegraphics[width=0.86\textwidth]{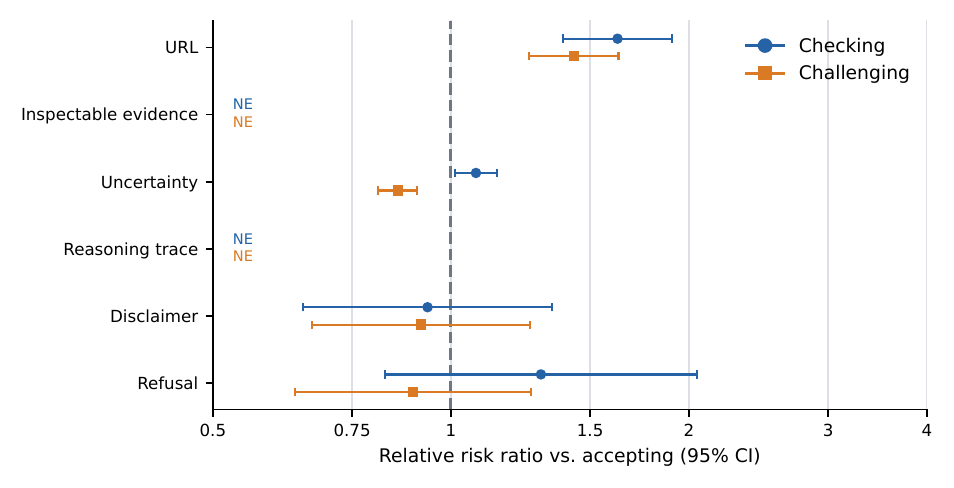}
\caption{Technical Help: uncertainty and the differing checking/challenging estimates.}
\label{fig:new_rq3_tech}
\end{figure}

Within Seeking Information, refusals and reasoning-trace indicators show stronger challenging associations (Figure~\ref{fig:new_rq3_seeking}).

\begin{figure}[H]
\centering
\includegraphics[width=0.86\textwidth]{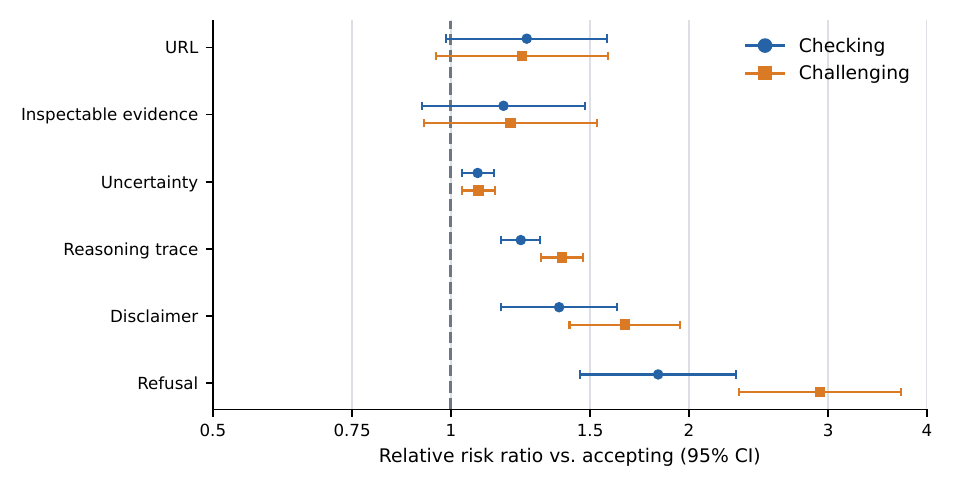}
\caption{Seeking Information: differential refusal and reasoning-trace associations.}
\label{fig:new_rq3_seeking}
\end{figure}

The conversation-grouped comparison favors initial task information over response features alone (Figure~\ref{fig:new_rq3_heldout}).

\begin{figure}[H]
\centering
\includegraphics[width=0.86\textwidth]{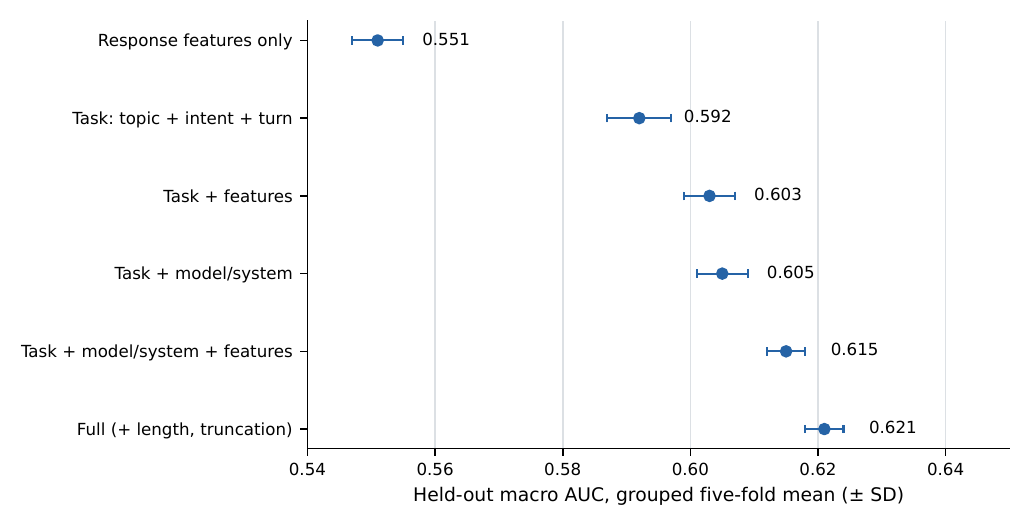}
\caption{Conversation-grouped predictive comparisons, with topic/intent/turn blocks and response-feature blocks.}
\label{fig:new_rq3_heldout}
\end{figure}

The pooled lag differs markedly from fixed-effects and bias-corrected estimates (Figure~\ref{fig:new_rq4_lag}).

\begin{figure}[H]
\centering
\includegraphics[width=0.86\textwidth]{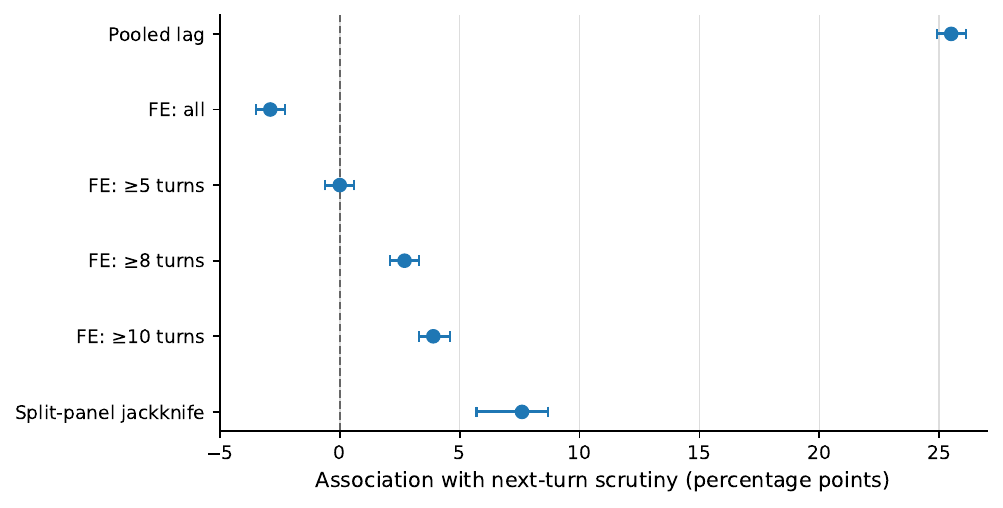}
\caption{Pooled and conversation-fixed-effect lag estimates and jackknife. This chart shows association, not causal carry-over.}
\label{fig:new_rq4_lag}
\end{figure}

The scrutiny share is relatively flat across normalized conversation position, as shown in Figure~\ref{fig:position_scrutiny}.

\begin{figure}[H]
\centering
\includegraphics[width=0.86\textwidth]{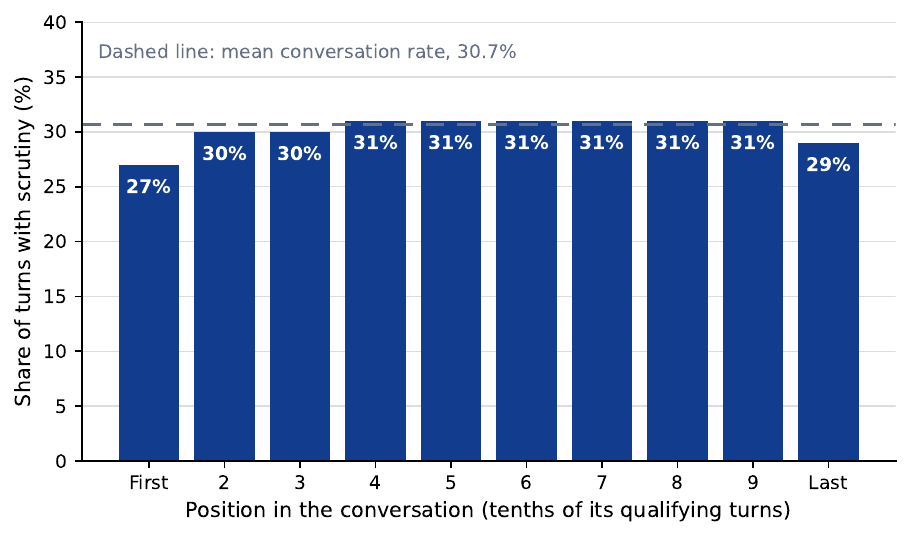}
\caption{Share of qualifying turns classified as scrutiny by normalized position within the conversation, in the at-least-five-turn sample.}
\label{fig:position_scrutiny}
\end{figure}

The exploratory reasoning-versus-nonreasoning comparison is shown in Figure~\ref{fig:reasoning_family}; task and deployment differences confound model-family assignment.

\begin{figure}[H]
\centering
\includegraphics[width=0.86\textwidth]{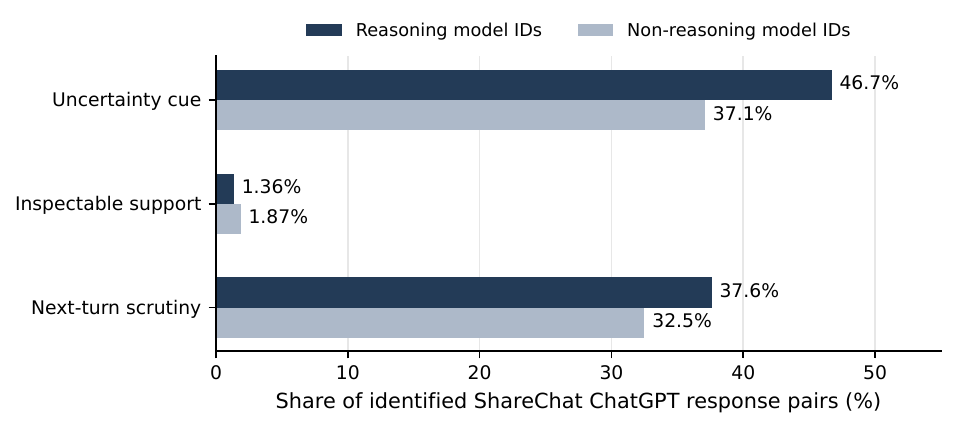}
\caption{Reasoning versus non-reasoning model family among model-identified ShareChat ChatGPT responses: uncertainty, inspectable support, and subsequent scrutiny. Descriptive, nonrandomized comparisons.}
\label{fig:reasoning_family}
\end{figure}

Two examples from the WildChat model-version comparison appear in Figure~\ref{fig:wildchat_versions}; they are not a complete model-version ranking.

\begin{figure}[H]
\centering
\includegraphics[width=0.86\textwidth]{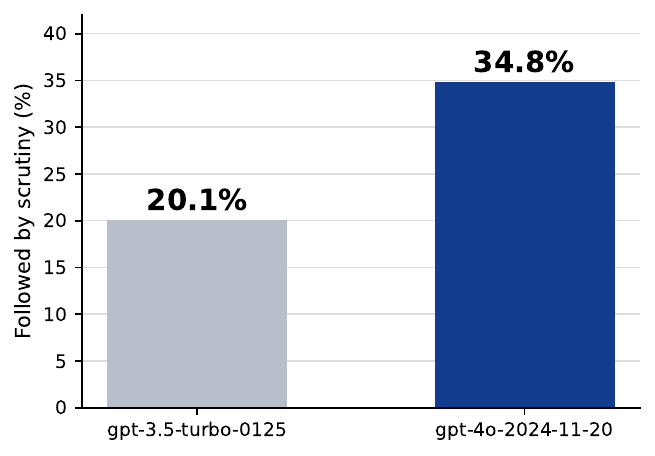}
\caption{Endpoint contrast for model-version scrutiny rates within WildChat; complete counts for other versions not shown in the source presentation.}
\label{fig:wildchat_versions}
\end{figure}

The unequal availability of preserved source fields is summarized in Figure~\ref{fig:sourcefields_deck}.

\begin{figure}[H]
\centering
\includegraphics[width=0.86\textwidth]{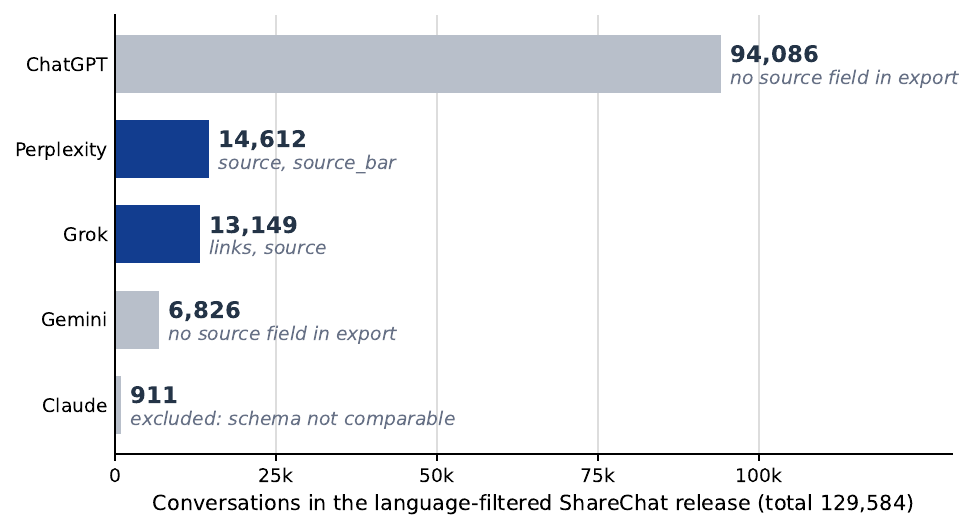}
\caption{ShareChat platform composition and available preserved source-object fields.}
\label{fig:sourcefields_deck}
\end{figure}

\end{document}